\documentclass[10pt,journal,compsoc]{IEEEtran}

\usepackage{amssymb}
\usepackage{amstext}
\usepackage{amsmath}
\usepackage{amsthm}
\usepackage{multicol, multirow}
\usepackage{graphicx}
\usepackage{dsfont}
\usepackage{tikz}
\usepackage{ctable}
\usepackage{pifont}
\usepackage{overpic}
\usepackage{circuitikz}
\usepackage{tikzscale}
\usepackage{pstricks}
\usepackage{pgfplots}
\newlength\figureheight 
\newlength\figurewidth
\usepackage{soul}
\usepackage{tabu}
\usepackage{url}
\usepackage{arydshln}
\usepackage{cite}
\usepackage{enumitem}

\newcommand{\cmark}{\ding{51}}

\DeclareMathDelimiter{\lVert}
  {\mathopen}{symbols}{"6B}{largesymbols}{"0D}
\DeclareMathDelimiter{\rVert}
  {\mathclose}{symbols}{"6B}{largesymbols}{"0D}

\usepackage{microtype}

\begin{document}

\begin{titlepage}
\null
\vfill
\renewcommand{\fboxsep}{10pt}
\centering
\fbox{\Large\begin{minipage}{\columnwidth}
\textbf{Disclaimer:}

This work has been accepted for publication in the IEEE Transactions on Pattern Analysis and Machine Intelligence:\vspace{4pt}
\newline
DOI: 10.1109/TPAMI.2022.3175371
\newline
\newline
\textbf{Copyright:} 
\newline
\copyright~2022 IEEE. Personal use of this material is permitted. Permission from IEEE must be obtained for all other uses, in any current or future media, including reprinting/republishing this material for advertising or promotional purposes, creating new collective works, for resale or redistribution to servers or lists, or reuse of any copyrighted component of this work in other works
\newline
\end{minipage}}
\vfill
\clearpage
\end{titlepage}

% https://www.ieee.org/content/dam/ieee-org/ieee/web/org/pubs/author_version_faq.pdf

\title{Learning general and distinctive 3D local\\deep descriptors for point cloud registration}

\author{Fabio Poiesi and Davide Boscaini% <-this % stops a space
\IEEEcompsocitemizethanks{\IEEEcompsocthanksitem Fabio Poiesi and Davide Boscaini are with the Technologies of Vision Lab in Fondazione Bruno Kessler, via Sommarive 18, Trento, 38123, Italy. E-mails: poiesi@fbk.eu, dboscaini@fbk.eu.
\protect
\IEEEcompsocthanksitem This research was supported by the SHIELD project, funded by the European Union’s Joint Programming Initiative – Cultural Heritage, Conservation, Protection and Use joint call, and partially by Provincia Autonoma di Trento (Italy) under L.P.~6/99  as part of the X-Loader4.0 project.
\protect
}
\thanks{Manuscript received mm dd, yyyy; revised mm dd, yyyy.}}

\markboth{Journal of \LaTeX\ Class Files,~Vol.~nn, No.~n, August~yyyy}%
{Shell \MakeLowercase{\textit{et al.}}: Bare Demo of IEEEtran.cls for Computer Society Journals}

\makeatletter
\long\def\@IEEEtitleabstractindextextbox#1{\parbox{0.922\textwidth}{#1}}
\makeatother

\IEEEtitleabstractindextext{
\begin{abstract}
An effective 3D descriptor should be invariant to different geometric transformations, such as scale and rotation, robust to occlusions and clutter, and capable of generalising to different application domains.
We present a simple yet effective method to learn general and distinctive 3D local descriptors that can be used to register point clouds that are captured in different domains.
Point cloud patches are extracted, canonicalised with respect to their local reference frame, and encoded into scale and rotation-invariant compact descriptors by a deep neural network that is invariant to permutations of the input points.
This design is what enables our descriptors to generalise across domains.
We evaluate and compare our descriptors with alternative handcrafted and deep learning-based descriptors on several indoor and outdoor datasets that are reconstructed by using both RGBD sensors and laser scanners.
Our descriptors outperform most recent descriptors by a large margin in terms of generalisation, and also become the state of the art in benchmarks where training and testing are performed in the same domain.
\end{abstract}
}

\maketitle
\IEEEdisplaynontitleabstractindextext
\IEEEpeerreviewmaketitle

%%%%%%%%%%%%%%%%%%%%%%%%%%%%%%%%%%%%%%%%%%%%%%%%%%%%%%%%%%%%%%
%%%%%%%%%%%%%%%%%%%%%%%%%%%%%%%%%%%%%%%%%%%%%%%%%%%%%%%%%%%%%%
%%%%%%%%%%%%%%%%%%%%%%%%%%%%%%%%%%%%%%%%%%%%%%%%%%%%%%%%%%%%%%
\IEEEraisesectionheading{\section{Introduction}\label{sec:introduction}}

3D point set registration is the problem of finding an optimal Euclidean transformation to align two partially overlapping 3D point sets such that they can be represented in a common reference frame.
3D point set registration is important for numerous applications including loop detection \cite{Zhou2022} and object 6D pose estimation \cite{Drost2010}.
3D point set registration approaches can estimate this transformation from 3D points directly or through geometrically-informed descriptors \cite{Rusu2009,Deng2018cvpr,Poiesi2021,Ao2021}.
In practice, there exists two main approaches.
Approaches that iteratively minimise the distance between nearest-neighbouring points if their original reference frames are loosely aligned, e.g.~Iterative Closest Point (ICP) \cite{Besl1992}, and approaches that find corresponding points through robust descriptor matching, e.g.~RANSAC \cite{Fischler1981}.
This paper focuses on the latter, in particular on the design of general and distinctive compact descriptors for 3D point clouds that are captured in the real world \cite{Rusu2009, Zeng2017, Deng2018cvpr, Choy2019, Bai2020, Poiesi2021, Ao2021}.
This problem is particularly challenging because point clouds can be captured with different sensors (e.g.~RGBD \cite{Zeng2017}, LiDAR \cite{Geiger2012}) and in different contexts (e.g.~indoors, outdoors), thus requiring descriptors to be effective across different domains.

Descriptors can be computed with or without a local reference frame (LRF) \cite{Salti2014,Yang2016,Gojcic2019,Poiesi2021,Choy2019,Bai2020,Ao2021}.
LRF is a rigid transformation that enables descriptor rotation-invariance \cite{Salti2014,Yang2016}. 
In turn, descriptors can be handcrafted or based on deep learning.
Deep learning-based descriptors \cite{Zeng2017,Deng2018cvpr,Deng2018eccv,Deng2019,Zhao2019,Gojcic2019,Choy2019,Bai2020,Poiesi2021,Ao2021} largely outperform their handcrafted counterpart \cite{Johnson1999,Rusu2009,Zhong2009,Salti2014,Yang2016}, and can also be computationally efficient if sparse convolutions are used~\cite{Choy2019}.
These efficient methods can compactly encode local geometric information for each point (dense local descriptors) \cite{Choy2019,Bai2020}, however they lack generalisation ability when trained and tested on different domains.
Differently, patch-based descriptors encode local geometric information for a given patch at chosen keypoints \cite{Gojcic2019,Poiesi2021,Ao2021}.
They have better generalisation ability than their dense counterpart, but reduced efficiency.

In this paper we present a novel deep learning-based method to learn General and Distinctive (GeDi) patch-based descriptors for point cloud registration.
Given a patch defined as a set of neighbouring points, we achieve rotation-invariance by computing a LRF that aims to rigidly transform the patch into its canonical representation \cite{Yang2016}.
We process this canonicalised patch using a point permutation-invariant deep network to produce the unitary-length descriptor of the patch \cite{Qi2017b}.
Our deep network uses receptive fields with different radii to aggregate local geometric information at multiple scales.
To mitigate the problem of a noisily estimated LRF, we learn a quaternion transformation network (QNet) that refines the canonicalisation operation before encoding the points into a descriptor.
We train QNet and the encoding network concurrently through a Siamese approach by using contrastive learning \cite{Choy2019}.
We perform an extensive evaluation to assess the generalisation and the distinctive abilities of GeDi on three large-scale datasets, i.e.~ETH \cite{Pomerleau2012}, KITTI \cite{Geiger2012} and 3DMatch \cite{Zeng2017}.
We compare our method with 19 methods from the literature and show that GeDi outperforms them by a large margin in terms of generalisation ability, while also being as effective as other descriptors when trained and tested on the same domain.
The source code is publicly available at \emph{\url{https://github.com/fabiopoiesi/gedi}}.

This paper extends our earlier work \cite{Poiesi2021} in several aspects.
We add QNet and modify the deep network architecture used in \cite{Poiesi2021} in order to process patches at multiple scales.
QNet ensures that the transformation applied to the canonicalised points lives in $SO(3)$, property that is not guaranteed with the spatial transformer network (TNet) \cite{Poiesi2021}, unless orthogonality is enforced, e.g., via a regularisation loss.
Without orthogonality, TNet may distort points, thus making multi-scale aggregation inconsistent across neighbourhoods with different sizes.
QNet can be trained without needing this regularisation loss or a dedicated Chamfer loss \cite{Zhao2019,Poiesi2021}.
We change our training strategy, moving from a farthest-point sampling of patches to a fully randomised strategy.
In \cite{Poiesi2021}, we processed the same points during training because we relied on data pre-processing as in \cite{Gojcic2019}.
In this new work we perform random sampling at each iteration for the points transformed with the LRF and also for the canonicalised points encoded by the deep network.
As a result, randomised sampling of the input point distribution acts as effective data augmentation.
We also provide a comprehensive review of related works.
Lastly, we significantly expand our experimental evaluation and analysis, considering additional recent methods, adding a new dataset (i.e.~KITTI odometry \cite{Geiger2012}), and including new qualitative and quantitative results.
%%%%%%%%%%%%%%%%%%%%%%%%%%%%%%%%%%%%%%%%%%%%%%%%%%%%%%%%%%%%%%%%%%%%%%
%%%%%%%%%%%%%%%%%%%%%%%%%%%%%%%%%%%%%%%%%%%%%%%%%%%%%%%%%%%%%%%%%%%%%%
%%%%%%%%%%%%%%%%%%%%%%%%%%%%%%%%%%%%%%%%%%%%%%%%%%%%%%%%%%%%%%%%%%%%%%
\section{Related work}\label{sec:soa}

Descriptors for point cloud registration can be designed by using handcrafted or deep learning-based algorithms, which in turn can be computed with or without local reference frame (LRF).
Typically, handcrafted methods encode local geometric structures as histograms of, e.g., point coordinates, surface normals, and/or pairwise point/normal relationships, whereas deep learning-based methods learn dense or sparse representations of local geometric structures.
Regardless of their formulation, these descriptors can be utilised with registration algorithms, such as RANSAC \cite{Fischler1981}.

\emph{Handcrafted methods without LRF} can encode local descriptors by using 3D point coordinates \cite{Cirujeda2014} and surface normals \cite{Johnson1999,Rusu2009,Drost2010}.
The former includes the MCOV descriptor that is computed directly from a covariance matrix encoding the correlation between the geometric and the photometric information of a patch \cite{Cirujeda2014}, such as relative angles and RGB values of points.
The authors of MCOV show that local geometric representations promote robustness to rigid spatial transformations, clutter and variations of point densities.
The latter includes descriptors such as SpinImage~\cite{Johnson1999}, Fast Point Feature Histograms (FPFH)~\cite{Rusu2009} and Point-Pair Features (PPF)~\cite{Drost2010}.
The SpinImage descriptor is based on rotation-invariant parametric representations of points within a patch~\cite{Johnson1999}.
% relative to the coordinate and the normal of its centroid
These representations are projected on a 2D image plane and mapped to a descriptor by computing a 2D histogram.
FPFH and PPF use features that are computed as distances and relative angles between points and normals.
FPFH is defined as the histogram of these features, while PPF is designed as point representations in the form of a hash table in order to make the task of point cloud registration efficient.
While FPFH can be used to register large point clouds, PPF is more suitable for object 6D pose estimation.

\emph{Handcrafted LRF-based methods} aim to compute a descriptor by firstly estimating a LRF, typically using the points of the patch, and secondly by building a compact representation that encodes the geometric properties of these points transformed by their LRF.
This category includes the Intrinsic Shape Signatures (ISS) descriptor that is computed as a weighted covariance matrix that aggregates the geometric properties of points within the patch.
The weights penalise distant points from the centroid.
The LRF is obtained through the eigenvalue decomposition of this covariance matrix \cite{Zhong2009}.
ISS is then built using the occupational histogram of a spherical neighbourhood with a predefined radius around the centroid.
The Signature of Histograms of Orientations (SHOT) descriptor defines the LRF as the eigenvalue decomposition of the covariance matrix to find two orthogonal axes \cite{Salti2014}.
Unlike ISS, SHOT disambiguates the sign of these axes based on the geometric structure of the patch.
The third axis is the cross product between these two axes.
SHOT is then built as a combination of an occupational histogram and of the relative geometric properties within the patch.
To improve robustness to clutter, the Triple Orthogonal Local Depth Images (TOLDI) descriptor extends SHOT's LRF by introducing a weighted aggregation of the points \cite{Yang2016}.
Points are projected on the three 2D planes defined by its LRF, and the TOLDI descriptor is then built as the 2D histograms of the projections.
Similary to ISS, SHOT and TOLDI, the Rotational Projection Statistics (RoPS) \cite{Guo2013} descriptor uses a covariance matrix to define its LRF, but based on the local connectivity of the points.
Points are then projected on the 2D planes defined by its LRF, and the descriptor is built using low-order central moments and entropy of these points.

\emph{Deep learning-based methods without LFR} model descriptor rotation invariance by either considering pairs of points, or by learning it via data augmentation, or by simply ignoring it.
The 3DMatch approach transforms patches into volumetric voxel grids of Truncated Distance Function (TDF) values and processes them through a 3D convolutional network to output local descriptors \cite{Zeng2017}.
The PPFNet approach adds normals and PPF representations \cite{Drost2010} of the patch to the point coordinates, and processes this information with a PointNet-based deep network \cite{Qi2017a}.
PPFNet aggregates local and global features to build context-informed descriptors \cite{Deng2018cvpr}.
PPF-FoldNet implements the idea of learning descriptors by means of an autoencoder \cite{Yang2018}, which is trained to reconstruct its input PPF representations.
PPF-FoldNet discards point coordinate and normal information, and learns to produce descriptors directly from the PPF representations without using supervision \cite{Deng2018eccv}.
PPF-FoldNet is based on PointNet.
The FCGF approach uses a fully-convolutional deep network to generate descriptors through 3D sparse convolutions \cite{Choy2019}.
Because sparse convolutions are efficient to compute, FCGF processes the whole point cloud in one pass and outputs a descriptor per point.
Unlike FCGF, D3Feat extracts descriptors by using dense deep features through a KPConv backbone \cite{Thomas2017,Bai2020}.
The SpinNet approach produces rotation invariant representations by using a spatial point transformer that project the input points to a cylindrical space \cite{Ao2021}. 
A deep network that is based on 3D cylindrical convolutional layers is then utilised to output patch-based descriptors.

\emph{Deep learning-based methods with LRF} aim to compute descriptors by training deep networks to process input points that are canonicalised with respect to their LRF.
The 3DSmoothNet approach uses TOLDI's LRF to canonicalise points \cite{Yang2016, Gojcic2019}.
As 3DMatch, 3DSmoothNet transforms points into volumetric voxel grids and processes them with a 3D convolutional network, however, instead of assigning TDF values to the voxels, 3DSmoothNet computes Gaussian smoothed representations based on the coordinates of the points within the patch.
As 3DSmoothNet, the distinctive local 3D descriptors (DIP) use TOLDI to canonicalise points \cite{Poiesi2021} and PointNet to produce local descriptors.
DIP uses a transformation network on the input points to improve eventual noisy canonicalisations.
The local multi-view descriptors (LMVD) are computed through multi-viewpoint image rendering through a differentiable renderer \cite{Li2020}.
For a point of interest, LMVD defines the LRF of the viewpoint by using the point normal and a consistent upright orientation.
LMVD produces descriptors by extracting local features maps from each viewpoint and by aggregating them via soft-view pooling.
Authors in \cite{Spezialetti2019} propose an unsupervised method to learn descriptors by using a Spherical CNN encoder that produces rotation-equivariant representations and a decoder that reconstructs the input points.
Authors show how the LRF can be learnt in a end-to-end manner by exploiting the Spherical CNN output as it lives in $SO(3)$.
%%%%%%%%%%%%%%%%%%%%%%%%%%%%%%%%%%%%%%%%%%%%%%%%%%%%%%%%%%%%%%%%%%%%%%
%%%%%%%%%%%%%%%%%%%%%%%%%%%%%%%%%%%%%%%%%%%%%%%%%%%%%%%%%%%%%%%%%%%%%%
%%%%%%%%%%%%%%%%%%%%%%%%%%%%%%%%%%%%%%%%%%%%%%%%%%%%%%%%%%%%%%%%%%%%%%
\section{Our approach}\label{sec:approach}

Let $\mathcal{X} = \{\mathbf{x}\} \subset \mathcal{P}$ be a local patch extracted from a point cloud $\mathcal{P} \subset \mathbb{R}^3$, where $\mathcal{X}$ is an unordered set of 3D points.
We design an algorithm that calculates a compact descriptor of $\mathcal{X}$ such that $\mathbf{f} = (\Phi_{\boldsymbol{\Theta}} \circ \Psi)(\mathcal{X})$, where 
$\mathbf{f} \in \mathbb{R}^d$ is the $d$-dimensional descriptor of $\mathcal{X}$, 
$\Psi$ is the function that samples and canonicalises $\mathcal{X}$ through LRF,
and $\Phi_{\boldsymbol{\Theta}}$ is a deep network with learnable parameters $\boldsymbol{\Theta}$.
Without loss of generality, we use the 3D coordinates of $\mathcal{X}$ to compute $\mathbf{f}$, i.e.~$\mathbf{x} = (x, y, z)$.

%%%%%%%%%%%%%%%%%%%%%%%%%%%%%%%%%%%%%%%%
%%%%%%%%%%%%%%%%%%%%%%%%%%%%%%%%%%%%%%%%
\subsection{Local sampling and canonicalisation}\label{sec:local_sampling_and_canon}

Sampling and canonicalisation involves (i) patch extraction, (ii) LRF estimation, and (iii) sampling of the points within the patch prior to processing them with the deep network.

Let $\mathcal{X} = \{\mathbf{x} : \lVert \mathbf{x} - \mathbf{\hat{x}} \rVert_2 \le r \}$ be the patch, where 
$\mathbf{\hat{x}} \in \mathcal{P}$ is the patch centre, 
$r$ is the patch radius and
$\lVert \cdot \rVert_2$ is the Euclidean norm.
Patches extracted from different regions of $\mathcal{P}$ may have different cardinalities due to occlusions or clutter.
To learn a general representation of patches having the same geometric structures but with different cardinalities, we randomly sample $m$ points within $\mathcal{X}$.
If patches have less than $m$ points, we pad to $m$ points by sampling with replacement from $\mathcal{X}$.
Patches with the same cardinality also enable us to efficiently train $\Phi_{\boldsymbol{\Theta}}$ using minibatches.
Let $S_m$ be the sampling function to produce the set of randomly sampled points $\mathcal{\tilde{X}} = S_m(\mathcal{X})$, where $\mathcal{\tilde{X}} \subset \mathcal{X}$ and $\lvert \mathcal{\tilde{X}} \rvert = m$.

The LRF estimation of $\mathcal{\tilde{X}}$ involves the computation of three orthogonal axes: an axis corresponding to the normal of a plane estimated from $\tilde{\mathcal{X}}$, an axis corresponding to a repeatable vector lying on this plane, and an axis that is orthogonal to the previous two axes.
Let $\mathbf{R}_\mathbf{\hat{x}}$ be the rotation matrix that we use to transform $\mathcal{\tilde{X}}$ from the reference frame of $\mathcal{P}$ to the LRF of $\mathcal{\tilde{X}}$.
We use TOLDI to compute $\mathbf{R}_\mathbf{\hat{x}}$ \cite{Yang2016}.

Before transforming $\mathcal{\tilde{X}}$ with respect to its LRF, we sample $n$ points from $\mathcal{\tilde{X}}$, where $n < m$.
Let $S_n$ be the sampling function to produce the set of randomly sampled points $\mathcal{\bar{X}} = S_n(\mathcal{\tilde{X}})$, where $\mathcal{\bar{X}} \subset \mathcal{\tilde{X}}$ and $\lvert \mathcal{\bar{X}} \rvert = n$.
To promote descriptor translation invariance, we represent the coordinates of $\mathcal{\bar{X}}$ relative to their patch centre $\mathbf{\hat{x}}$.
To promote descriptor scale invariance, we normalise these coordinates relative to $\mathbf{\hat{x}}$ by the radius of the patch $r$,
%++++++++++++++++++++
\begin{equation}
\Psi(\mathcal{\bar{X}}) = \{\mathbf{y} : \mathbf{y} = \mathbf{R}_\mathbf{\hat{x}} ((\mathbf{x} - \mathbf{\hat{x}}) / r), \mathbf{x} \in \mathcal{\bar{X}} \},
\end{equation}
%++++++++++++++++++++
which becomes the input to our deep network.

Note that $n < m$ to speed up the deep network computation, to handle large batches during training and because we experimentally observed that distinctiveness is upper-bounded when a certain number of points is reached.

%%%%%%%%%%%%%%%%%%%%%%%%%%%%%%%%%%%%%%%%
%%%%%%%%%%%%%%%%%%%%%%%%%%%%%%%%%%%%%%%%
\subsection{Deep network design}\label{sec:deep_net_design}

Because $\Psi(\mathcal{\bar{X}})$ is an unordered set of points, we use a deep network that is robust to permutations of the input points~\cite{Qi2017b}.
Like DIP~\cite{Poiesi2021}, we use a L2 normalisation layer to transform the network output into a $d$-dimensional descriptor $\mathbf{f}$ with unitary norm.
Unlike DIP \cite{Poiesi2021}, which is computed using a single receptive field for the whole set of input points~\cite{Qi2017a}, our deep network uses a hierarchical structure of the receptive fields to build high-dimensional representations from local geometric structures at multiple scales~\cite{Qi2017b}.
Our deep network learns to aggregate and to encode these representations using different kernel sizes along the hierarchy, making our descriptor more general and distinctive than DIP.
Specifically, to aggregate and encode local geometric representations, we use a PointNet++ deep network~\cite{Qi2017b}.
Because these multi-scale receptive fields cannot be used along with the spatial transformer network used in DIP, we use a different spatial transformer network that outputs a quaternion instead of a $3\times3$ transformation.

% ********************************
% ********************************
\begin{figure*}[t]
  \centering
  \includegraphics[width=1\textwidth]{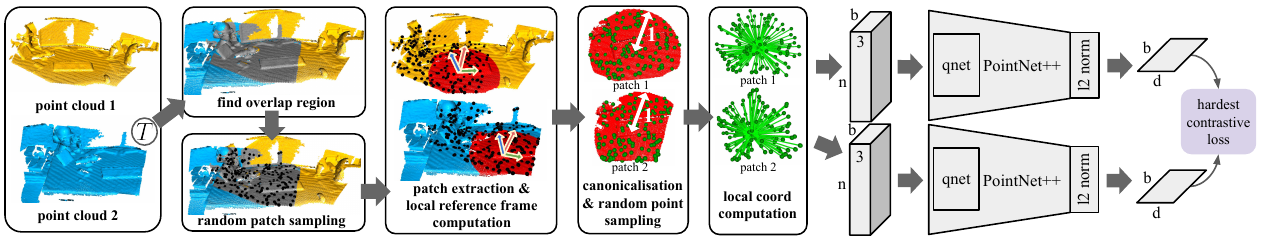}
  \vspace{-.7cm}
  \caption{
  GeDi's training pipeline.
  Two overlapping point clouds are aligned using a ground-truth transformation ($T$).
  From the overlap region (grey) we randomly sample $b$ patch centroids (black).
  We use a Siamese approach to train two deep neural networks with shared parameters concurrently.
  Prior to the deep network processing, we perform the following operations:
  (i) for each centre (black) a patch (red) with radius $r$ is extracted and the corresponding local reference frame is computed using a subset of the points of the patch;
  (ii) this patch is canonicalised using the local reference frame and $n$ points are randomly sampled from the patch (green points);
  (iii) the coordinates of these $n$ points are represented relative to the patch centre and normalised to obtain patches with unitary radius;
  (iv) these $n$ points are given to the deep networks as input to learn the descriptor.
  The L2-normalisation layer is used to output descriptors with unitary norm.
  We train the deep network using the hardest-contrastive loss \cite{Choy2019}.
  }
  \label{fig:training_block_diagram}
\end{figure*}
% ********************************
% ********************************

%%%%%%%%%%%%%%%%%%%%%%%%%%%%%%%%%%%%%%%%
%%%%%%%%%%%%%%%%%%%%%%%%%%%%%%%%%%%%%%%%
\noindent\textbf{Quaternion network.}
DIP uses a dedicated deep network (TNet~\cite{Qi2017a}) to estimate a transformation that can mitigate eventual noisy canonicalisations of the input points.
Because PointNet++'s grouping layers operate in the metric space through a ball query operation with fixed radius \cite{Qi2017b}, TNet's transformation should live in $SO(3)$.
TNet's output can be constrained to be a $SO(3)$ transformation via the regularisation loss term $\ell_{\mathrm{reg}}$=$\lVert \mathbf{I} - \mathbf{A}\mathbf{A}^\top \rVert^2_F$, where $\mathbf{A} \in \mathbb{R}^{3\times3}$ is TNet's output \cite{Qi2017a}.
Alternatively, SVD orthogonalisation can be applied to TNet's output to produce a $SO(3)$ transformation.
We experimentally observed that these two solutions are suitable for PointNet++ and lead to similar results, however the former needs an extra loss term and weight hyperparameter, while the latter is less efficient due to the SVD orthogonalisation step.
Therefore, inspired by \cite{Yuan2019}, we introduce a PointNet-based deep network, namely QNet, which outputs a unit-norm quaternion, hence a $SO(3)$ transformation by construction.
QNet uses a L2 normalisation layer at the PoinNet's output to produce the unit-norm vector.
QNet can be learnt concurrently with PointNet++, and neither requires an additional loss term with a weight hyperparameter to tune, nor adds computational overhead.

%%%%%%%%%%%%%%%%%%%%%%%%%%%%%%%%%%%%%%%%
%%%%%%%%%%%%%%%%%%%%%%%%%%%%%%%%%%%%%%%%
\noindent\textbf{Training procedure.}
Descriptors should be similar between corresponding patches and dissimilar between non-corresponding patches of different point clouds.
We train our network through a Siamese approach that processes pairs of corresponding patches by using two branches with shared weights \cite{Deng2018cvpr,Gojcic2019}.
Fig.~\ref{fig:training_block_diagram} shows our training architecture.

Training begins by selecting point cloud pairs that overlap at least $\tau_o$ \cite{Gojcic2019}.
We randomly sample $b$ pairs of corresponding points from the overlap region, which we treat as our patch centres.
We perform local sampling and canonicalisation for each patch (Sec.~\ref{sec:local_sampling_and_canon}), and input the result of this to our Siamese network with shared parameters.
Each branch calculates a descriptor for a given patch.
We learn the network parameters using a hardest-contrastive loss \cite{Choy2019}, where the hardest negatives are sampled within each minibatch.
There exist different strategies to mine negatives.
DIP uses a FPS-based approach to avoid selecting negatives that are spatially close to the anchors \cite{Poiesi2021}.
In GeDi, we mine negatives outside a spherical region of radius $r_\mathcal{C}$ centred on each anchor \cite{Choy2019}.
We experimentally found that this mining strategy combined with the random sampling of patches can be implemented efficiently, and enable us to build larger minibatches while promoting randomness during training.

%%%%%%%%%%%%%%%%%%%%%%%%%%%%%%%%%%%%%%%%%%%%%%%%%%%%%%%%%%%%%%%%%%%%%%
%%%%%%%%%%%%%%%%%%%%%%%%%%%%%%%%%%%%%%%%%%%%%%%%%%%%%%%%%%%%%%%%%%%%%%
%%%%%%%%%%%%%%%%%%%%%%%%%%%%%%%%%%%%%%%%%%%%%%%%%%%%%%%%%%%%%%%%%%%%%%
\section{Experiments}\label{sec:experiments}

We assess the generalisation ability of GeDi descriptors by using three large-scale point cloud datasets: 3DMatch \cite{Zeng2017}, ETH \cite{Pomerleau2012}, and KITTI \cite{Geiger2012}. 
We follow the same train/test splits as in \cite{Ao2021}.
3DMatch is composed of 62 indoor scenes that are captured with RGBD sensors, containing dense point clouds of reconstructed rooms.
The ground-truth transformations are from \cite{Gojcic2019}.
As in \cite{Gojcic2019}, we also use an augmented version of 3DMatch, namely 3DMatchR, where each point cloud is randomly rotated by an angle within $[0^{\circ}, 360^{\circ}]$.
ETH includes four outdoor scenes that are captured with a laser scanner, containing partially overlapping point clouds \cite{Gojcic2019}.
These sequences are used for testing only and the ground-truth transformations are from \cite{Gojcic2019}.
KITTI contains 11 sequences of outdoor driving scenarios that are captured using a laser scanner.
We follow the same evaluation setup as in \cite{Choy2019}.
We compare GeDi with other descriptors using the same evaluation procedures as in \cite{Ao2021}.
We compare GeDi against: 
SpinImage~\cite{Johnson1999}, 
FPFH~\cite{Rusu2009}, 
USC~\cite{Tombari2010}, 
SHOT~\cite{Salti2014},
3DMatch~\cite{Zeng2017},
CGF~\cite{Khoury2017},
FoldingNet~\cite{Yang2018},
3DFeat-Net~\cite{Yew2018},
PPFNet~\cite{Deng2018cvpr},
PPF-FoldNet~\cite{Deng2018eccv},
DirectReg~\cite{Deng2019},
CapsuleNet~\cite{Zhao2019},
Equivariant3D~\cite{Spezialetti2019},
PerfectMatch~\cite{Gojcic2019},
FCGF~\cite{Choy2019},
D3Feat~\cite{Bai2020} (we will use \emph{-rand} and \emph{-pred} as suffixes when descriptors are chosen randomly and through D3Feat's keypoint detector, respectively),
LMVD~\cite{Li2020},
SpinNet~\cite{Ao2021} and
DIP~\cite{Poiesi2021}.

Our experiments are organised as follows: 
(i) we train GeDi on 3DMatch, and test it on ETH and KITTI; 
(ii) we train GeDi on KITTI and test it on 3DMatch;
(ii) we train and test GeDi on data of the same dasasets according to their respective original train/test splits;
(iv) we measure the inference time and perform an ablation study of GeDi.

%%%%%%%%%%%%%%%%%%%%%%%%%%%%%%%%%%%%%%%%%%%%%%%%%%%%%%%%%%%%%%%%%%%%%%
%%%%%%%%%%%%%%%%%%%%%%%%%%%%%%%%%%%%%%%%%%%%%%%%%%%%%%%%%%%%%%%%%%%%%%
\subsection{Evaluation measures}

For 3DMatch and ETH, we use the feature-matching recall (FMR) \cite{Deng2018cvpr} to quantify the descriptor distinctiveness and to compare our results with other approaches.
We use the same FMR's notation as in \cite{Poiesi2021}: $\Xi$ indicates FMR, $\mu_\xi$ and $\sigma_\xi$ indicate the mean and standard deviation of the average number of correctly matched patch pairs over the point clouds of the dataset, respectively.
FMR indirectly measures the number of iterations required by a registration algorithm, e.g.~RANSAC, to estimate the transformation between two point clouds~\cite{Deng2018eccv}.
For KITTI, we randomly select a set of points from each point cloud pair, compute their descriptors, and use RANSAC \cite{Fischler1981} to estimate the rigid transformation to register them.
We set RANSAC as in~\cite{Yew2018}.
The estimated rigid transformation is compared to its ground truth by using the Relative Translational Error (RTE) and Relative Rotation Error (RRE)~\cite{Ma2016}.
We compute the success rate by considering a registration to be successful when both RTE and RRE are less then 2m and $5^{\circ}$, respectively.
We report the average RTE and RRE values of the successful cases.
Unlike RTE and RRE, FMR does not require RANSAC as it directly averages the number of correctly matched point clouds across datasets.

%%%%%%%%%%%%%%%%%%%%%%%%%%%%%%%%%%%%%%%%%%%%%%%%%%%%%%%%%%%%%%%%%%%%%%
%%%%%%%%%%%%%%%%%%%%%%%%%%%%%%%%%%%%%%%%%%%%%%%%%%%%%%%%%%%%%%%%%%%%%%
\subsection{Parameters}\label{sec:parameters}

\textbf{Training}. We randomly sample $b$=$350$ patch centres in the overlap region between each point cloud pair.
In 3DMatch we sample $m$=$4000$ points for the LRF computation, while we use $m$=$2000$ in KITTI because its point clouds are sparser than those of 3DMatch.
We experimentally observed that higher values of $m$ have minor effects on the results.
After the LRF computation we sample $n$=$512$ points per patch to train our deep network.
The descriptor dimension is always set to $d$=$32$ as in \cite{Ao2021}.
We set $r$=$.5$m for 3DMatch and $r$=$2.5$m for KITTI.
For 3DMatch, we train for 10 epochs, performing 16K iterations per epoch.
Each iteration processes a point cloud pair.
For KITTI, we train for 122 epochs, performing 1.4K iterations per epoch.
We use Stochastic Gradient Descent with an initial learning rate of $10^{-1}$ that decreases by a factor $.1$ every 3 epochs in 3DMatch and every 35 epochs in KITTI.
We set weight decay to 5$\cdot$$10^{-5}$ and Dropout to $.3$ at the last MLP layer.
In the hardest-contrastive loss, we set the same margins as in \cite{Choy2019} and the radius for negative mining equal to $.2r$.
Because LRFs may slightly differ across different subsampled versions of the same patch, we learn descriptors that are robust to these variations by randomly rotating canonicalised patches during training.
We rotate by independently sampling three angles in the range $[-10^{\circ}$,$10^{\circ}]$.

\noindent\textbf{Testing}. We set $n$=$1024$ as we found it works well in practice.
Patch radii are set to $r$=$.6$m in 3DMatch, $r$=$1.5$m in ETH and $r$=$2.5$m in KITTI.
Sec.~\ref{sec:ablation_study} reports our ablation study for $n$ and $r$.
As in \cite{Gojcic2019}, in 3DMatch and ETH, we compute the descriptors for $c$=5K randomly sampled points from each point cloud and compute their FMR scores.
KITTI is typically tested by computing the descriptor of each point and by estimating the rigid transformation between two point clouds through RANSAC \cite{Ao2021}.
KITTI point clouds are about 120K+ points, and the computation of a GeDi descriptor for each point is lengthy.
Therefore, we randomly sample $c$=5K, 25K and 50K points from each point cloud, and compare our results against the other approaches.
This setup is more challenging than the original one as it requires a greater discriminative ability of the descriptors as fewer are used.

%%%%%%%%%%%%%%%%%%%%%%%%%%%%%%%%%%%%%%%%%%%%%%%%%%%%%%%%%%%%%%%%%%%%%%
%%%%%%%%%%%%%%%%%%%%%%%%%%%%%%%%%%%%%%%%%%%%%%%%%%%%%%%%%%%%%%%%%%%%%%
\subsection{Quantitative analysis}
\label{sec:quantitative_analysis}

%%%%%%%%%%%%%%%%%%%%%%%%%%%%%%%%%%%%%%%%%%%%%%%%%%%%%%%%%%%%%%%%%%%%%%
\noindent\textbf{RGBD indoors $\pmb{\rightarrow}$ laser scanner outdoors.}
Tab.~\ref{tab:3dm2eth_res} shows the results of GeDi trained on 3DMatch and tested on ETH.
The results of FPFH, SHOT, 3DMatch, CGF and PerfectMatch are taken from \cite{Gojcic2019}.
The results of FCGF and D3Feat are taken from \cite{Bai2020}.
The results of the other methods are taken from the respective papers.
GeDi outperforms the other methods by a large margin, improving FMR by 5.4\% with respect to DIP~\cite{Poiesi2021} and SpinNet \cite{Ao2021}.
Tab.~\ref{tab:3dm2kitti_res} shows the results of GeDi trained on 3DMatch and tested on KITTI.
These results are reproduced by us, using authors' source codes and the parameters authors used in this setup.
Where possible, we used subsets of points for the evaluation.
GeDi consistently outperforms the other methods by a large margin, achieving a success rate of +5.41\% and +7.93\% compared to DIP and SpinNet, respectively, with $c$=50K.
With $c$=5K, GeDi achieves +7.92\% and +17.65\% compared to DIP and SpinNet, respectively.

The experiments in Tabs.~\ref{tab:3dm2eth_res} and \ref{tab:3dm2kitti_res} show that deep learning patch-based descriptors (i.e.~PerfectMatch, DIP, SpinNet, GeDi) are the best performing ones, whereas deep learning dense local descriptors, i.e.~FCGF and D3Feat, show weaker generalisation ability.
Unlike the other methods, D3Feat-pred uses a keypoint detector to predict the most likely points that can provide informative descriptors.
Processing entire point clouds in a single pass to produce dense descriptors appears to affect invariance to rotations and to hinder the descriptor generalisation ability on different domains.
PerfectMatch, LMVD and SpinNet compute handcrafted input data representations, and process these with 2D or 3D convolutional deep networks: PerfectMatch uses smoothed density value voxelization \cite{Gojcic2019}, LMVD uses multi-view projections \cite{Li2020} and SpinNet uses spherical voxelisation \cite{Ao2021}.
Although input data representations are designed either to retain most of the original 3D information or to enable invariance to geometric transformations, we can observe that these methods have an overall weaker generalisation ability than architectures that process 3D points directly, like DIP and GeDi.
Unlike DIP, GeDi uses PointNet++ to make patch representations more informative through a cascade of receptive fields with different sizes, enabling us to capture and aggregate local geometric features at different scales.

%+++++++++++++++++++++++++
\begin{table}[t]
    \tabcolsep 2pt
    \centering
    \caption[]{Feature-matching recall ($\Xi$): 3DMatch (training) $\rightarrow$ ETH (testing).\\Keys: H: Handcrafted. S: Supervised.}
    \vspace{-.3cm}
    \label{tab:3dm2eth_res}
    \resizebox{1\columnwidth}{!}{%
    \begin{tabular}{lcccccc}
        \toprule
        \multirow{3}{*}{Method} & \multirow{3}{*}{Type} & \multicolumn{2}{c}{Gazebo} & \multicolumn{2}{c}{Wood} & \\ 
        & & Summer & Winter & Autumn & Summer \\
        & & $\Xi$ {\footnotesize$\uparrow$} & $\Xi$ {\footnotesize$\uparrow$} & $\Xi$ {\footnotesize$\uparrow$} & $\Xi$ {\footnotesize$\uparrow$} & avg {\footnotesize$\uparrow$} \\
        \midrule
        FPFH~\cite{Rusu2009}            & H & .386 & .142 & .148 & .208 & .221 \\
        SHOT~\cite{Salti2014}           & H & .739 & .457 & .609 & .640 & .611 \\
        3DMatch~\cite{Zeng2017}         & S & .228 & .083 & .139 & .224 & .169 \\
        CGF~\cite{Khoury2017}           & S & .375 & .138 & .104 & .192 & .202 \\
        PerfectMatch~\cite{Gojcic2019}  & S & .913 & .841 & .678 & .728 & .790 \\
        FCGF~\cite{Choy2019}            & S & .228 & .100 & .148 & .168 & .161 \\
        D3Feat-rand~\cite{Bai2020}      & S & .457 & .239 & .130 & .224 & .262 \\
        D3Feat-pred~\cite{Bai2020}      & S & .859 & .630 & .496 & .480 & .616 \\
        LMVD~\cite{Li2020}              & S & .853 & .720 & .840 & .783 & .799 \\
        DIP~\cite{Poiesi2021}           & S & .908 & .886 & .965 & .952 & .928 \\
        SpinNet~\cite{Ao2021}           & S & .929 & .917 & .922 & .944 & .928 \\
        \vspace{-.25cm} & & & & & & \\
        \hdashline[1pt/3pt]
        \vspace{-.25cm} & & & & & & \\
        GeDi                            & S & \textbf{.989} & \textbf{.965} & \textbf{.974} & \textbf{1.000} & \textbf{.982} \\
        \bottomrule
    \end{tabular}
    }
\end{table}
%+++++++++++++++++++++++++

%+++++++++++++++++++++++++
\begin{table}[t]
    \tabcolsep 5pt
    \centering
    \caption[]{Relative Translational Error (RTE) and Relative Rotation Error (RRE): 3DMatch (training) $\rightarrow$ KITTI (testing).}
    \vspace{-.3cm}
    \label{tab:3dm2kitti_res}
    \resizebox{1\columnwidth}{!}{%
    \begin{tabular}{llccccc}
        \toprule
        \multirow{2}{*}{$c$} & \multirow{2}{*}{Method} & \multicolumn{2}{c}{RTE[cm]} & \multicolumn{2}{c}{RRE[$^\circ$]} & \multirow{2}{*}{Suc.[\%]{\footnotesize$\uparrow$}} \\
        & & avg {\footnotesize$\downarrow$} & std {\footnotesize$\downarrow$} & avg {\footnotesize$\downarrow$} & std {\footnotesize$\downarrow$} & \\
        \midrule
        \multirow{6}{*}{\rotatebox[origin=c]{90}{{\centering 5K}}} & FCGF~\cite{Choy2019} & 31.21 & 2.87 & 1.60 & 1.15 & 34.95 \\
        & D3Feat-rand~\cite{Bai2020} & 41.39 & 3.48 & 1.68 & 1.25 & 20.72 \\
        & D3Feat-pred~\cite{Bai2020} & 33.66 & 3.12 & 1.50 & 1.18 & 38.74 \\
        & SpinNet~\cite{Ao2021} & 16.67 & 1.64 & 1.13 & 0.92 & 65.41 \\
        & DIP~\cite{Poiesi2021} & 13.58 & 1.25 & 0.93 & 0.92 & 75.14 \\
        \vspace{-.25cm} & & & & & & \\
        \cdashline{2-7}[1pt/3pt]
        \vspace{-.25cm} & & & & & & \\
        & GeDi & 10.34 & 0.70 & 0.40 & 0.49 & 83.06 \\
        \midrule
        \multirow{6}{*}{\rotatebox[origin=c]{90}{{\centering 25K}}} & FCGF~\cite{Choy2019} & 28.18 & 2.46 & 1.63 & 1.19 & 44.68 \\
        & D3Feat-rand~\cite{Bai2020} & 34.23 & 2.87 & 1.54 & 1.19 & 24.50 \\
        & D3Feat-pred~\cite{Bai2020} & 35.67 & 3.22 & 1.61 & 1.25 & 22.16 \\
        & SpinNet~\cite{Ao2021} & 12.53 & 1.76 & 0.76 & 0.68 & 86.67 \\
        & DIP~\cite{Poiesi2021} & 9.84 & 0.85 & 0.56 & 0.61 & 91.17 \\
        \vspace{-.25cm} & & & & & & \\
        \cdashline{2-7}[1pt/3pt]
        \vspace{-.25cm} & & & & & & \\
        & GeDi & 8.43 & 0.75 & 0.40 & 0.46 & 97.30 \\
        \midrule
        \multirow{2}{*}{\rotatebox[origin=c]{90}{{\centering all}}} & FCGF~\cite{Choy2019} & 25.93 & 2.45 & 1.52 & 1.14 & 45.77 \\
        & D3Feat~\cite{Bai2020} & 37.71 & 3.25 & 1.54 & 1.22 & 25.77 \\
        \midrule
        \multirow{3}{*}{\rotatebox[origin=c]{90}{{\centering 50K}}} & SpinNet~\cite{Ao2021} & 12.03 & 1.11 & 0.68 & 0.70 & 90.99 \\
        & DIP~\cite{Poiesi2021} & 9.45 & 0.73 & 0.48 & 0.59 & 93.51 \\
        \vspace{-.25cm} & & & & & & \\
        \cdashline{2-7}[1pt/3pt]
        \vspace{-.25cm} & & & & & & \\
        & GeDi & 8.21 & 0.70 & 0.40 & 0.51 & 98.92 \\
        \bottomrule
    \end{tabular}
    }
\end{table}
%+++++++++++++++++++++++++

%%%%%%%%%%%%%%%%%%%%%%%%%%%%%%%%%%%%%%%%%%%%%%%%%%%%%%%%%%%%%%%%%%%%%%
\noindent\textbf{Laser scanner outdoors $\pmb{\rightarrow}$ RGBD indoors.}
Tab.~\ref{tab:kitti23dm_res} confirms GeDi's generalisation ability across different sensors and contexts, i.e.~training on KITTI and testing on 3DMatch.
The results of FCGF, D3Feat and SpinNet are taken from~\cite{Ao2021}.
FCGF and D3Feat show weaker generalisation ability than SpinNet and GeDi, and a significant drop in FMR on 3DMatchR, suggesting that they cannot effectively encode general rotation-invariant properties.
GeDi outperforms SpinNet by 7.7\% both on the original and on the augmented version of 3DMatch, reinforcing that GeDi's design to produce descriptors from raw 3D points promotes generalisation.

%+++++++++++++++++++++++++
\begin{table}[t]
    \tabcolsep 7pt
    \centering
    \caption[]{Feature-matching recall ($\Xi$): KITTI (training) $\rightarrow$ 3DMatch (testing).}
    \vspace{-.3cm}
    \label{tab:kitti23dm_res}
    \resizebox{.8\columnwidth}{!}{%
    \begin{tabular}{lccccc}
        \toprule
        \multirow{2}{*}{Method} & \multicolumn{2}{c}{3DMatch} & \multicolumn{2}{c}{3DMatchR} \\
        & $\Xi$ {\footnotesize$\uparrow$} & std {\footnotesize$\downarrow$} & $\Xi$ {\footnotesize$\uparrow$} & std {\footnotesize$\downarrow$} \\
        \toprule
        FCGF~\cite{Choy2019}                & .325 & .074 & .010 & .010 \\
        D3Feat-rand~\cite{Bai2020}          & .607 & .077 & .172 & .046 \\
        D3Feat-pred~\cite{Bai2020}          & .627 & .081 & .178 & .032 \\
        SpinNet~\cite{Ao2021}               & .845 & \textbf{.059} & .842 & \textbf{.058} \\
        \vspace{-.25cm} & & & & \\
        \hdashline[1pt/3pt]
        \vspace{-.25cm} & & & & \\
        GeDi                                & \textbf{.922} & \textbf{.059} & \textbf{.919} & .059 \\
        \bottomrule
    \end{tabular}%
    }
\end{table}
%+++++++++++++++++++++++++

\begin{figure}[t]
\begin{center}
  \begin{tabular}{@{}c@{}c}
      \includegraphics[width=.5\columnwidth]{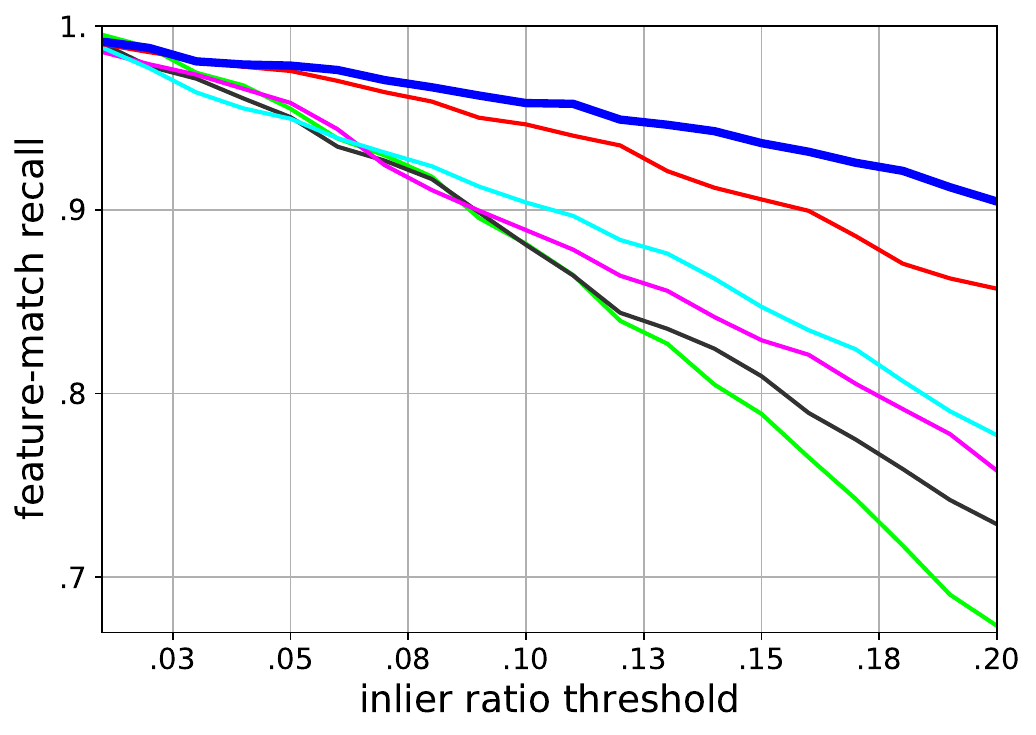} &
      \includegraphics[width=.5\columnwidth]{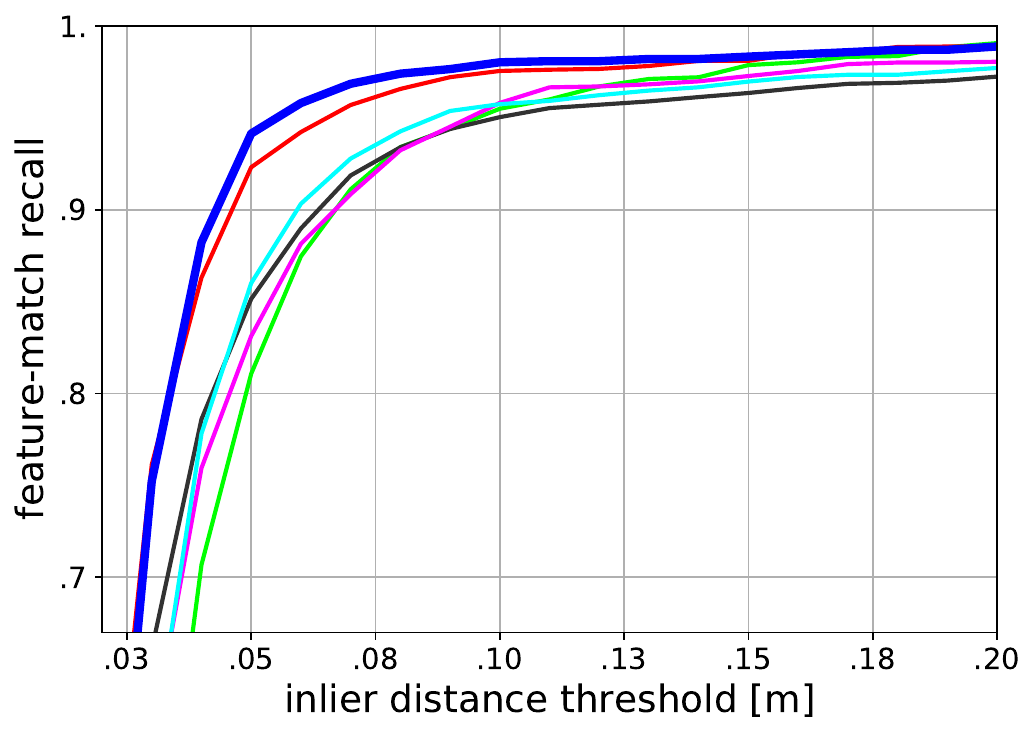}
      \vspace{-.15cm}
  \end{tabular}
  \begin{tabular}{@{}c}
    \includegraphics[width=.95\columnwidth]{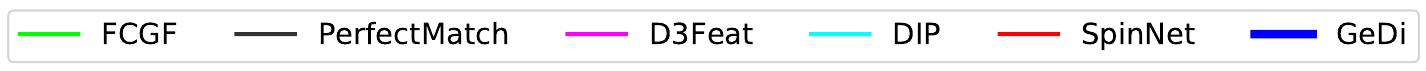}
  \end{tabular}
\end{center}
\vspace{-.5cm}
\caption{Feature-matching recall as a function (left) of the inlier-ratio and (right) of the inlier-distance thresholds \cite{Poiesi2021}}
\label{fig:inlier_graphs}
\end{figure}

%%%%%%%%%%%%%%%%%%%%%%%%%%%%%%%%%%%%%%%%%%%%%%%%%%%%%%%%%%%%%%%%%%%%%%
\noindent\textbf{Same settings.}
Tab.~\ref{tab:fmr_3dmatch} reports the results of GeDi on 3DMatch.
The results are taken from the following sources: 
SpinImage, SHOT, USC, FPFH and Equivariant3D are from \cite{Spezialetti2019};
CGF, 3DMatch and PPFNet are from \cite{Deng2018cvpr};
FoldingNet and PPF-FoldNet are from \cite{Deng2018eccv};
the other results are from the respective papers.
GeDi outperforms all the other descriptors on both 3DMatch and 3DMatchR, thus showing its better distinctiveness ability also on data of the same domain.
Fig.~\ref{fig:inlier_graphs} provides a detailed picture of the results on 3DMatch by varying FMR's inlier ratio and inlier distance thresholds \cite{Ao2021}.
We can observe that GeDi can produce a larger number of distinctive descriptors than SpinNet, especially when the inlier threshold is .20 we have a +4.8\% of FMR.
Tab.~\ref{tab:kitti2kitti} reports the results of GeDi on KITTI.
The results are taken from the original papers.
GeDi achieves a success rate of 99.82\%; only one pair out of 555 failed the geometric registration.
The success rate of GeDi is similar to that of D3Feat.
D3Feat's RTE and RRE are .65cm and $9^{\circ}$ lower than GeDi, respectively.
Although D3Feat's translational and rotational errors are lower, we deem GeDi's result comparable to that of D3Feat as the transformation that is estimated with RANSAC can be improved, e.g., by refining it with ICP.

%+++++++++++++++++++++++++
\begin{table}[t]
    \tabcolsep 4pt
    \caption[]{Feature-matching recall ($\Xi$): 3DMatch (training) $\rightarrow$ 3DMatch (testing). 
    Keys: U: Unsupervised. S: Supervised. H: Handcrafted.}
    \vspace{-.3cm}
    \label{tab:fmr_3dmatch}
    \resizebox{\columnwidth}{!}{%
    \begin{tabular}{llccccccc}
        \toprule
        \multirow{2}{*}{$c$} & \multirow{2}{*}{Method} & \multirow{2}{*}{Type} & \multicolumn{2}{c}{3DMatch} & \multicolumn{2}{c}{3DMatchR} & Feat.\\
        & & & $\Xi$ {\footnotesize$\uparrow$}  & std {\footnotesize$\downarrow$}  & $\Xi$ {\footnotesize$\uparrow$}  & std {\footnotesize$\downarrow$}  & dim. {\footnotesize$\downarrow$} \\
        \toprule
        \multirow{5}{*}{\rotatebox[origin=c]{90}{{\centering 2048}}} & FoldingNet~\cite{Yang2018} & U & .613 & .087 & .023 & .010 & 512 \\
        & PPFNet~\cite{Deng2018cvpr} & S & .623 & .108 & .003 & .005 & 64 \\
        & DirectReg~\cite{Deng2019} & S & .746 & .094 & - & - & 512 \\
        & CapsuleNet~\cite{Zhao2019} & U & .787 & .084 & .807 & .062 & 512\\
        \vspace{-.25cm} & & & & & & & \\
        \cdashline{2-8}[1pt/3pt]
        \vspace{-.25cm} & & & & & & & \\
        & GeDi & S & .979 & .022 & .977 & .025 & 32 \\
        \midrule
        \multirow{17}{*}{\rotatebox[origin=c]{90}{{\centering 5000}}} & CGF~\cite{Khoury2017} & S & .478 & .094 & .499 & .092 & 32 \\
        & 3DMatch~\cite{Zeng2017} & S & .596 & .073 & .011 & .010 & 512 \\
        & SpinImage~\cite{Johnson1999} & H & .633 & .106 & .639 & .098 & 153 \\
        & PPF-FoldNet~\cite{Deng2018eccv} & U & .718 & .105 & .731 & .104 & 512 \\
        & FPFH~\cite{Rusu2009} & H &.754 & .071 & .767 & .075 & 33 \\
        & USC~\cite{Tombari2010} & H & .868 & .052 & .877 & .053 & 1980 \\
        & SHOT~\cite{Salti2014} & H & .875 & .034 & .875 & .036 & 352 \\
        & Equivariant3D~\cite{Spezialetti2019} & U & .942 & .040 & .939 & .048 & 512 \\
        & PerfectMatch~\cite{Gojcic2019} & S & .947 & .027 & .949 & .025 & 32  \\
        & DIP~\cite{Poiesi2021} & S & .948 & .046 & .946 & .046 & 32 \\
        & FCGF~\cite{Choy2019} & S & .952 & .029 & .953 & .033 & 32  \\
        & D3Feat-rand~\cite{Bai2020} & S & .953 & .027 & .952 & .032 & 32 \\ 
        & D3Feat-pred~\cite{Bai2020} & S & .958 & .029 & .955 & .035 & 32 \\
        & LMVD~\cite{Li2020} & S & .975 & .028 & .969 & - & 32 \\
        & SpinNet~\cite{Ao2021} & S & .976 & \textbf{.019} & .975 & \textbf{.019} & 32 \\
        \vspace{-.25cm} & & & & & & & \\
        \cdashline{2-8}[1pt/3pt]
        \vspace{-.25cm} & & & & & & & \\
        & GeDi & S & \textbf{.979} & .022 & \textbf{.976} & .027 & 32 \\
        \bottomrule
    \end{tabular}%
    }
\end{table}
%+++++++++++++++++++++++++

% ********************************
% ********************************
% ********************************
% ********************************

%+++++++++++++++++++++++++
\begin{table}[t]
    \tabcolsep 5pt
    \centering
    \caption[]{Relative Translational Error (RTE) and Relative Rotation Error (RRE): KITTI (training) $\rightarrow$ KITTI (testing).}
    \vspace{-.3cm}
    \label{tab:kitti2kitti}
    \resizebox{.9\columnwidth}{!}{%
    \begin{tabular}{lcccccc}
        \toprule
        \multirow{2}{*}{Method} & \multicolumn{2}{c}{RTE[cm]} & \multicolumn{2}{c}{RRE[$^\circ$]} & \multirow{2}{*}{Suc.[\%]{\footnotesize$\uparrow$}} \\
        & avg {\footnotesize$\downarrow$} & std {\footnotesize$\downarrow$} & avg {\footnotesize$\downarrow$} & std {\footnotesize$\downarrow$} & \\
        \toprule
        3DFeat-Net~\cite{Yew2018}   & 25.9 & 26.2 & .57 & .46 & 95.97 \\
        FCGF~\cite{Choy2019}        & 9.52 & 1.30 & .30 & .28 & 96.57 \\
        DIP (25K pts)~\cite{Poiesi2021}       & 8.69 & .69 & .44 & .49 & 97.30 \\
        SpinNet~\cite{Ao2021}       & 9.88 & .50 & .47 & .09 & 99.10 \\
        % D3Feat-rand~\cite{Bai2020}  & 8.78 & .44 & .32 & .07 & 99.81 \\
        D3Feat~\cite{Bai2020}  & \textbf{6.90} & \textbf{.30} & \textbf{.24} & \textbf{.06} & 99.81 \\
        \vspace{-.25cm} & & & & & \\
        \hdashline[1pt/3pt]
        \vspace{-.25cm} & & & & & \\
        GeDi (25K pts)                  & 7.22 & .63 & .32 & .27 & 99.46 \\
        GeDi (50K pts)                  & 7.55 & .67 & .33 & .31 & \textbf{99.82} \\
        \bottomrule
    \end{tabular}
    }
\end{table}
%+++++++++++++++++++++++++

%%%%%%%%%%%%%%%%%%%%%%%%%%%%%%%%%%%%%%%%%%%%%%%%%%%%%%%%%%%%%%%%%%%%%%
\noindent\textbf{Inference time.}
We compare the inference time of different descriptors that are described in Sec.~\ref{sec:soa}.
Experiments are executed using an Intel Xeon CPU E5-1620v3 @ 3.50GHz with NVIDIA GTX 1070 8GB.
The chosen methods are FPFH~\cite{Rusu2009} (PCL implementation using OpenMP), SHOT~\cite{Salti2014} (PCL implementation using OpenMP), FCGF~\cite{Choy2019} and SpinNet~\cite{Ao2021}.
We use a point cloud from 3DMatch composed of about 41K points and a patch radius of .6m for the local descriptors.
FCGF requires point cloud voxelisation: we use a 2.5cm voxel size, which reduces the number of points processed by the deep network to about 16.5K.
We measured the inference time per descriptor and obtained 1.598ms for FPFH, 5.544ms for SHOT, 0.003ms for FCGF, 13.534ms for SpinNet and 1.454ms for GeDi.
FCGF resulted to be the fastest method, which is due to its efficient implementation based on the Minkowski Engine.

%%%%%%%%%%%%%%%%%%%%%%%%%%%%%%%%%%%%%%%%%%%%%%%%%%%%%%%%%%%%%%%%%%%%%%
%%%%%%%%%%%%%%%%%%%%%%%%%%%%%%%%%%%%%%%%%%%%%%%%%%%%%%%%%%%%%%%%%%%%%%
\subsection{Ablation study}\label{sec:ablation_study}

Tab.~\ref{tab:ablation_sampling_3dm} compares GeDi to other descriptors by reducing the number of sampled points from 5000 to 250.
The results of the other methods are taken from \cite{Ao2021}.
GeDi outperforms SpinNet by .9\% on average, and by 3\% when 250 points are sampled.
According to the FMR formulation, we can observe that if we randomly pick 250 points, 97.3\% of them on average are mutually-nearest neighbours in the descriptor space and close to each other in the metric space.
Therefore, the likelihood that RANSAC would successfully register two point clouds of this test set is rather high.

%+++++++++++++++++++++++++
\begin{table}[t]
    \tabcolsep 6pt
    \centering
    \caption{Ablation study using the feature-matching recall ($\Xi$) as a function of the number of sampled points on the 3DMatch dataset \cite{Zeng2017}.}
    \vspace{-.3cm}
    \label{tab:ablation_sampling_3dm}
    % \resizebox{1\columnwidth}{!}{%
    \begin{tabular}{lcccccc}
        \toprule
        \multirow{3}{*}{Method}         &  \multicolumn{5}{c}{\# sampled points} & \\
        & 5000 & 2500 & 1000 & 500 & 250 & \\
        & $\Xi$ {\footnotesize$\uparrow$} & $\Xi$ {\footnotesize$\uparrow$} & $\Xi$ {\footnotesize$\uparrow$} & $\Xi$ {\footnotesize$\uparrow$} & $\Xi$ {\footnotesize$\uparrow$} & avg {\footnotesize$\uparrow$} \\
        \toprule
        PerfectMatch~\cite{Gojcic2019}  & .947 & .942 & .926 & .901 & .829 & .909 \\
        FCGF~\cite{Choy2019}            & .952 & .955 & .946 & .930 & .899 & .936 \\
        D3Feat-rand~\cite{Bai2020}      & .953 & .951 & .942 & .936 & .908 & .938 \\
        D3Feat-pred~\cite{Bai2020}      & .958 & .956 & .946 & .943 & .933 & .947 \\
        SpinNet~\cite{Ao2021}           & .976 & .975 & .973 & .963 & .943 & .966 \\
        \vspace{-.25cm} & & & & & & \\
        \cdashline{1-7}[1pt/3pt]
        \vspace{-.25cm} & & & & & & \\
        GeDi                            & \textbf{.979} & \textbf{.977} & \textbf{.976} & \textbf{.972} & \textbf{.973} & \textbf{.975} \\
        \bottomrule
    \end{tabular}
    % }
\end{table}
%+++++++++++++++++++++++++

%+++++++++++++++++++++++++
\begin{table}[t]
    \tabcolsep 3pt
    \caption[]{Ablation study on different GeDi's implementation settings.\\Keys: ST: Spatial Transformer network. LRF: Local Reference Frame.\\The last row includes our setting in all the other experiments.}
    \vspace{-.3cm}
    \label{tab:abl_study}
    \resizebox{1\columnwidth}{!}{%
    \begin{tabular}{ccccccccccc}
        \toprule
        \multirow{2}{*}{$d$} & \multirow{2}{*}{$n$} & \multirow{2}{*}{$r$[m]} & \multirow{2}{*}{ST} & \multirow{2}{*}{LRF} & \multicolumn{3}{c}{3DMatch} & \multicolumn{3}{c}{3DMatchR} \\
        & & & & & $\Xi${\footnotesize$\uparrow$} & $\mu_\xi${\footnotesize$\uparrow$} & $\sigma_\xi${\footnotesize$\downarrow$} & $\Xi${\footnotesize$\uparrow$} & $\mu_\xi${\footnotesize$\uparrow$} & $\sigma_\xi${\footnotesize$\downarrow$} \\
        \toprule  
        32 & 1024 & .6 & & \cmark & .917 & .254 & .182 & .919 & .255 & .182 \\
        32 & 1024 & .6 & qnet & & .787 & .206 & .191 & .019 & .007 & .018 \\
        \midrule
        16 & 1024 & .6 & qnet & \cmark & .902 & .222 & .164 & .905 & .224 & .164 \\
        64 & 1024 & .6 & qnet & \cmark & .937 & .268 & .186 & .934 & .269 & .186 \\
        128 & 1024 & .6 & qnet & \cmark & .937 & .277 & .189 & .942 & .277 & .190 \\
        \midrule
        32 & 256 & .6 & qnet & \cmark & .903 & .206 & .146 & .896 & .207 & .146 \\
        32 & 512 & .6 & qnet & \cmark & .928 & .252 & .176 & .926 & .253 & .176 \\
        32 & 1536 & .6 & qnet & \cmark & .934 & .269 & .187 & .934 & .268 & .187 \\
        32 & 2048 & .6 & qnet & \cmark & .937 & .269 & .187 & .935 & .269 & .187 \\
        \midrule
        32 & 1024 & .2 & qnet & \cmark & .741 & .128 & .112 & .741 & .128 & .113 \\
        32 & 1024 & 1.0 & qnet & \cmark & .766 & .178 & .166 & .763 & .179 & .167 \\
        32 & 1024 & 1.4 & qnet & \cmark & .510 & .100 & .121 & .506 & .101 & .122 \\
        32 & 1024 & 1.8 & qnet & \cmark & .315 & .059 & .084 & .306 & .058 & .085 \\
        \midrule
        32 & 1024 & .6 & tnet+reg & \cmark & .932 & .265 & .183 & .930 & .266 & .184 \\
        32 & 1024 & .6 & tnet+svd & \cmark & .934 & .268 & .186 & .939 & .269 & .187 \\
        32 & 1024 & .6 & qnet & \cmark & .932 & .267 & .185 & .931 & .268 & .186 \\
        \bottomrule
    \end{tabular}
    }
\end{table}
%+++++++++++++++++++++++++

Next, we analyse a selection of GeDi configurations by using a subset of 3DMatch's training data.
We use \emph{7-scenes-chess} and \emph{7-scenes-fire} for training, and \emph{7-scenes-office} and \emph{7-scenes-pumpkin} for validation.
There are 2492 point cloud pairs for training and 1640 for validation.
In validation, we randomly sample 5K points per point cloud pair, for a total of 8.2M descriptors.
We train with the same parameters that are described in Sec.~\ref{sec:parameters}.
Tab.~\ref{tab:abl_study} reports the results on all our experimental configurations.
The last row is the configuration we used throughout all our previous experiments.

\begin{figure}[t]
\begin{center}
  \begin{tabular}{@{}c@{}c@{}c}
    \begin{overpic}[width=.33\columnwidth]{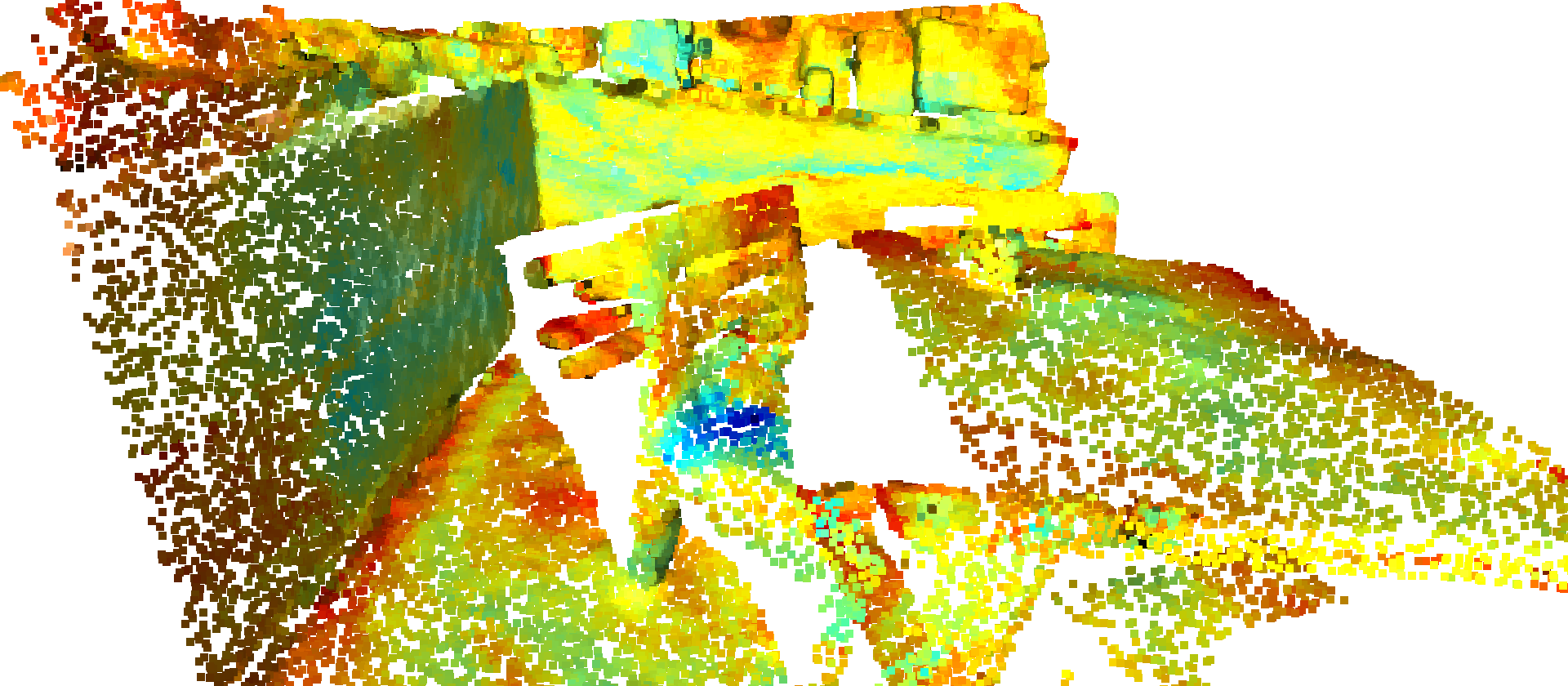}
      \put(3,-8){\color{black}\scriptsize\textbf{iter: 5, avg dist: 0.034}}
    \end{overpic}&
    \begin{overpic}[width=.33\columnwidth]{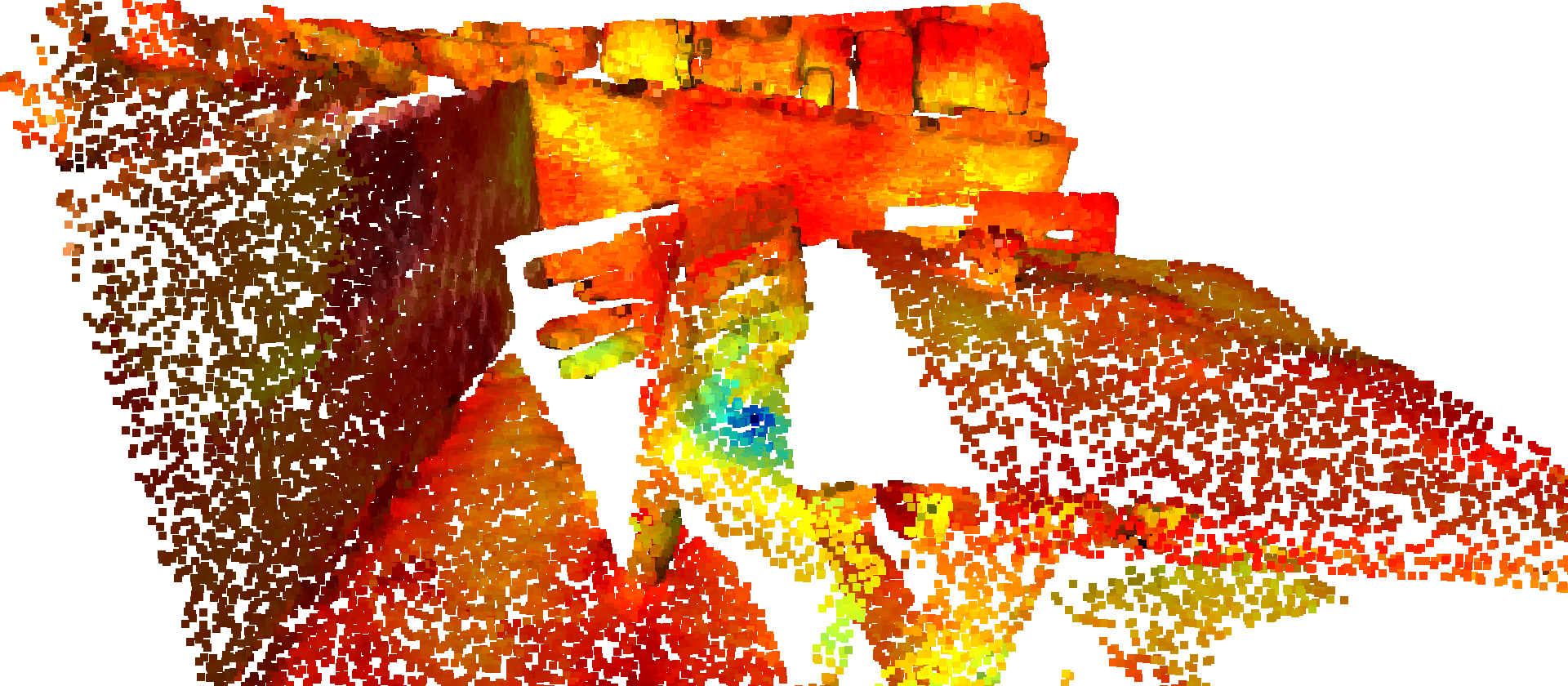}
      \put(3,-8){\color{black}\scriptsize\textbf{iter: 1500, avg dist: 0.950}}
    \end{overpic}&
    \begin{overpic}[width=.33\columnwidth]{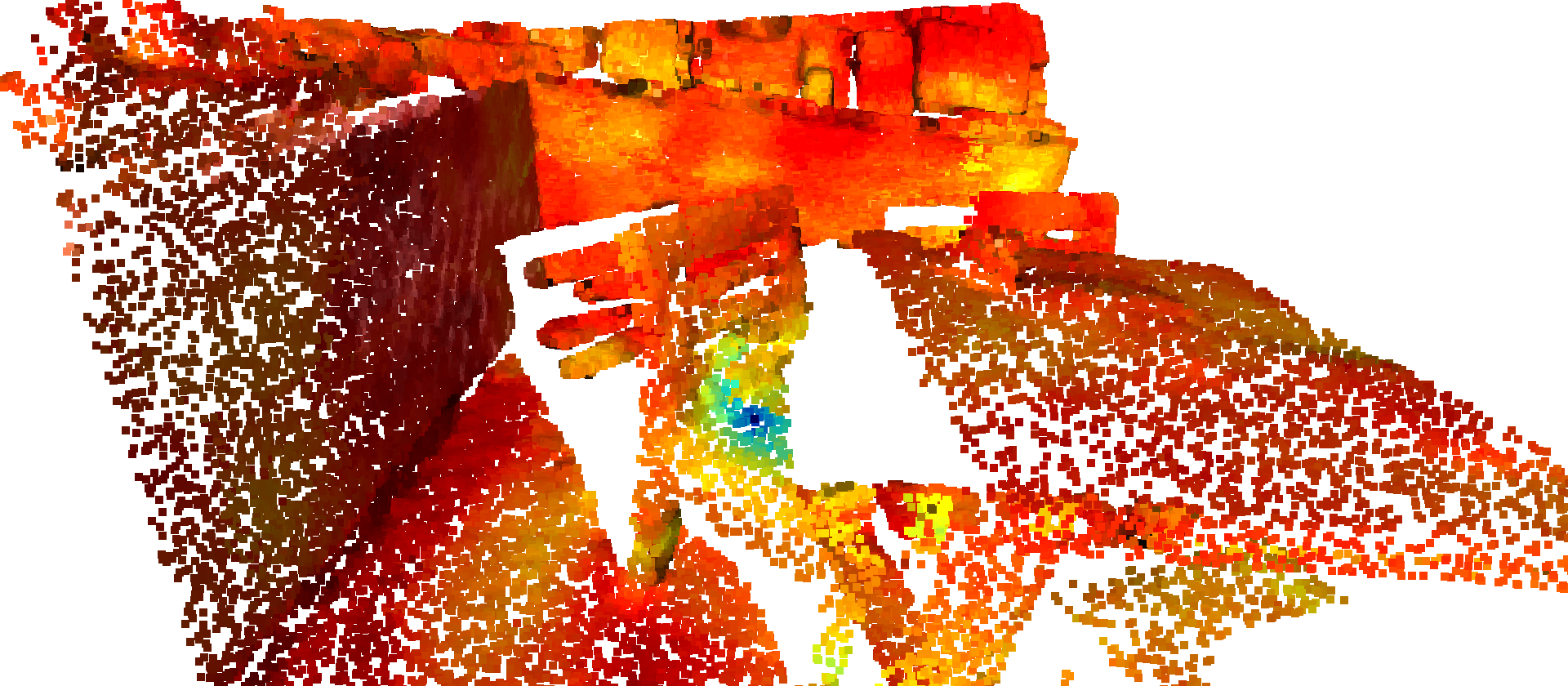}
      \put(3,-8){\color{black}\scriptsize\textbf{iter: 13450, avg dist: 1.031}}
    \end{overpic}\vspace{.3cm}\\
    \begin{overpic}[width=.33\columnwidth]{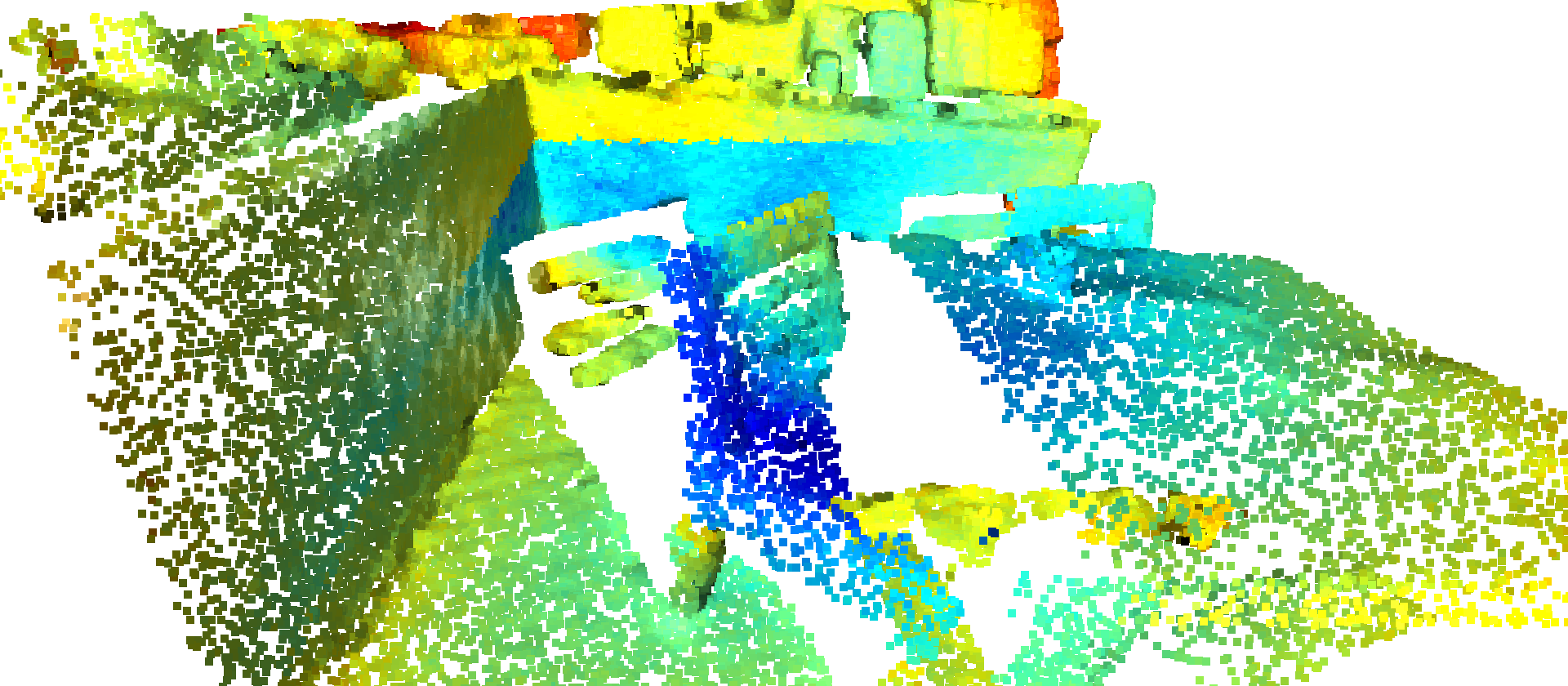}
      \put(3,-8){\color{black}\scriptsize\textbf{iter: 5, avg dist: 0.025}}
    \end{overpic}&
    \begin{overpic}[width=.33\columnwidth]{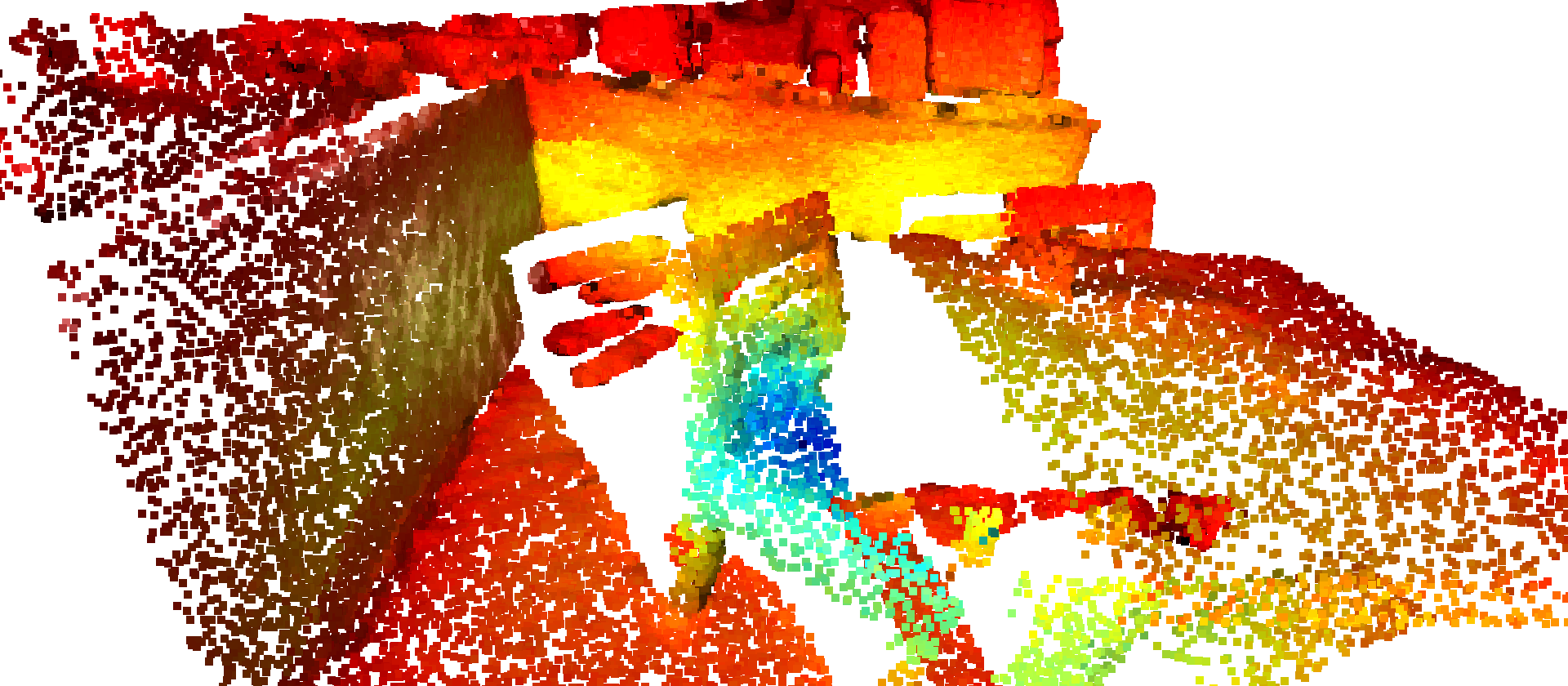}
      \put(3,-8){\color{black}\scriptsize\textbf{iter: 1500, avg dist: 0.902}}
    \end{overpic}&
    \begin{overpic}[width=.33\columnwidth]{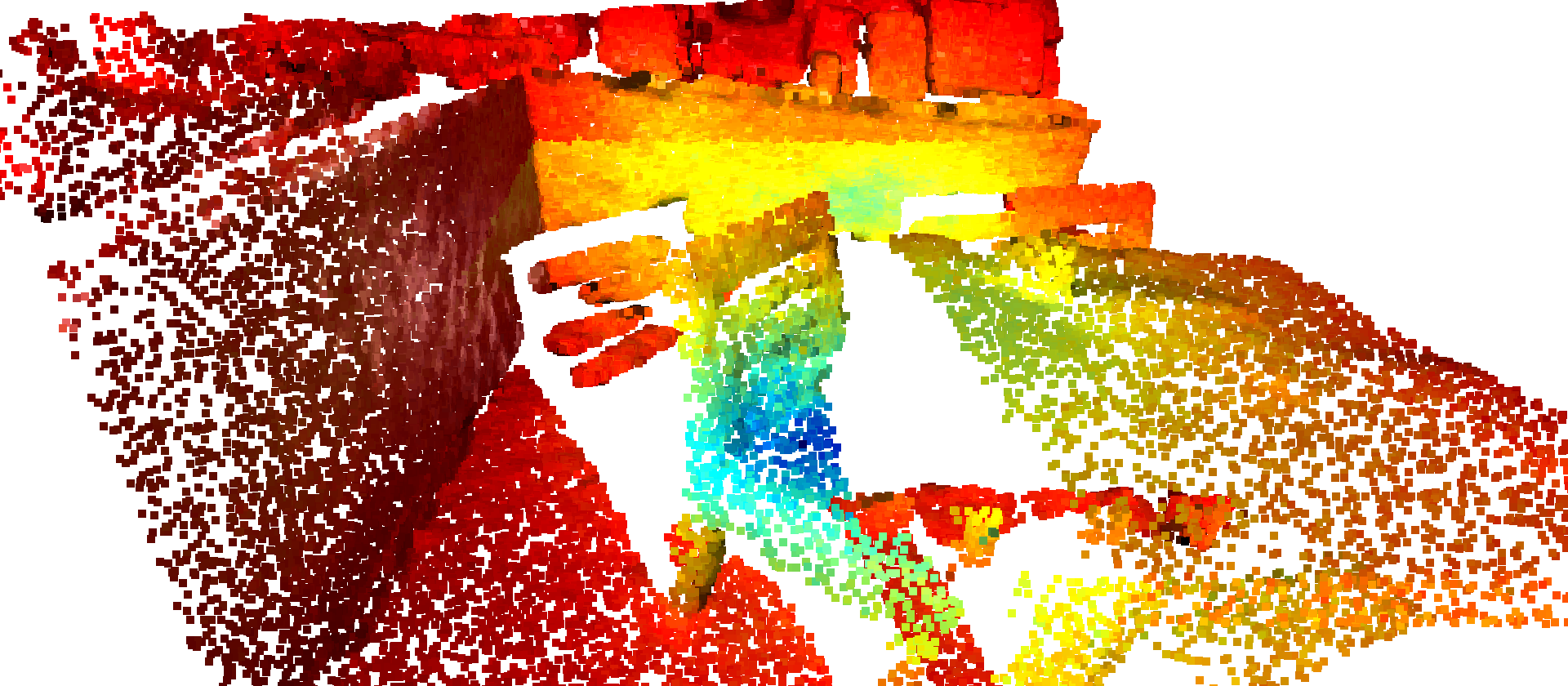}
      \put(3,-8){\color{black}\scriptsize\textbf{iter: 13450, avg dist: 0.959}}
    \end{overpic}
  \end{tabular}
\end{center}
\vspace{-.1cm}
\caption{Colour-coded distances (red=large, blue=small) between a descriptor and all the other descriptors.
The queried descriptor is located on the chair's seat in the centre of the image.
We used descriptor radii equal to .6 (top) and 3 (bottom).
Each figure reports the training iteration (iter) and the average distance with respect to all the other descriptors.}
\label{fig:descriptor_distance_map}
\end{figure}

The first part of Tab.~\ref{tab:abl_study} shows the usefulness of QNet and LRF canonicalisation.
GeDi without QNet provides worse results than the version with it.
To further investigate this behaviour, we compared the RRE between the rotations estimated by the LRFs and the composition of the latter with the rotations provided by QNet.
We analysed $\sim$1.4M matching patch pairs and Tab.~\ref{tab:chamfer_dist_qnet} includes the results.
We can observe that the RRE is smaller after the QNet estimated rotation is applied.
%+++++++++++++++++++++++++
\begin{table}[t]
    \centering
    \caption[]{RRE statistics before and after QNet computed on the ablation dataset. 
    About 1.4M matching patch pairs are used in this analysis.}
    \vspace{-.3cm}
    \label{tab:chamfer_dist_qnet}
    \resizebox{.6\columnwidth}{!}{%
    \begin{tabular}{lccc}
        \toprule
        & \textbf{median} & \textbf{mean} & \textbf{std} \\
        \midrule
        before QNet & 0.909 & 1.318 & 1.103 \\
        after QNet & 0.497 & 1.087 & 1.137 \\
        \bottomrule
    \end{tabular}
    }
\end{table}
%+++++++++++++++++++++++++
LRF is key to achieve descriptor rotation invariance as FMR significantly worsens in 3DMatchR.
The second part of Tab.~\ref{tab:abl_study} shows the effect on FMR when $d$ changes: we trained and evaluated GeDi using $d$=16, 64 and 128.
FMR improves as $d$ increases, suggesting that the larger the capacity to encode descriptor information, the better the performance.
The third part of Tab.~\ref{tab:abl_study} shows GeDi's behaviour when the number of points sampled from each patch changes before passing them to the deep network.
FMR improves as the number of points increases.
Although the deep network is trained using 512 points as input, it can also handle larger numbers of input points, at the cost of a higher inference time.
We used $n$=1024 throughout our experiments to have feedback in reasonable amounts of time.
In the fourth part of Tab.~\ref{tab:abl_study} we can see that a larger amount of contextual information that is encoded within the descriptor affects distinctiveness.
We analyse this behaviour by monitoring the distinctiveness throughout training, and by comparing cases when descriptors are learned using patch radii equal to .6 and 3.
We computed the Euclidean distance between a chosen descriptor and all the other descriptors, and colour-coded the point cloud based on this distance.
Fig.~\ref{fig:descriptor_distance_map} shows some examples of this when the radius is .6 (top) and 3 (bottom).
The top row shows a rather localised region of low distances, i.e.~only few descriptors in the neighbourhood of the chosen descriptor have a low distance.
The bottom row shows that a larger amount of contextual information encoded in the descriptor affects distinctiveness.
The patch radius also affects training time: the descriptor with radius .6 provides better distinctiveness with fewer training iterations than with radius 3.
In the last part of Tab.~\ref{tab:abl_study} we compare the results between QNet and TNet with enforced orthogonality.
We enforce TNet's output orthogonality by using (i) the regularisation loss $\ell_{\mathrm{reg}}$ (Sec.~\ref{sec:deep_net_design}) with weight .1 during training and (ii) the SVD orthogonalisation.
We name the former \emph{tnet+reg} and the latter \emph{tnet+svd}.
Results show that these three solutions are nearly equivalent in terms of FMR.
However, as opposed to \emph{tnet+reg}, we deem the solution with QNet more practical as it does not require neither an additional loss term nor a weighting hyperparameter to tune.
We found the solution with QNet computationally more efficient than \emph{tnet+svd} -- we measured a 1.33-time increase when the SVD orthogonalisation step is used.

% ********************************
% ********************************

%%%%%%%%%%%%%%%%%%%%%%%%%%%%%%%%%%%%%%%%%%%%%%%%%%%%%%%%%%%%%%%%%%%%%%
%%%%%%%%%%%%%%%%%%%%%%%%%%%%%%%%%%%%%%%%%%%%%%%%%%%%%%%%%%%%%%%%%%%%%%
\subsection{Qualitative results}

\begin{figure}[t]
\vspace*{3mm}
\begin{center}
  \begin{tabular}{@{}c@{\,}c}
    % correct
    \begin{overpic}[width=.48\columnwidth]{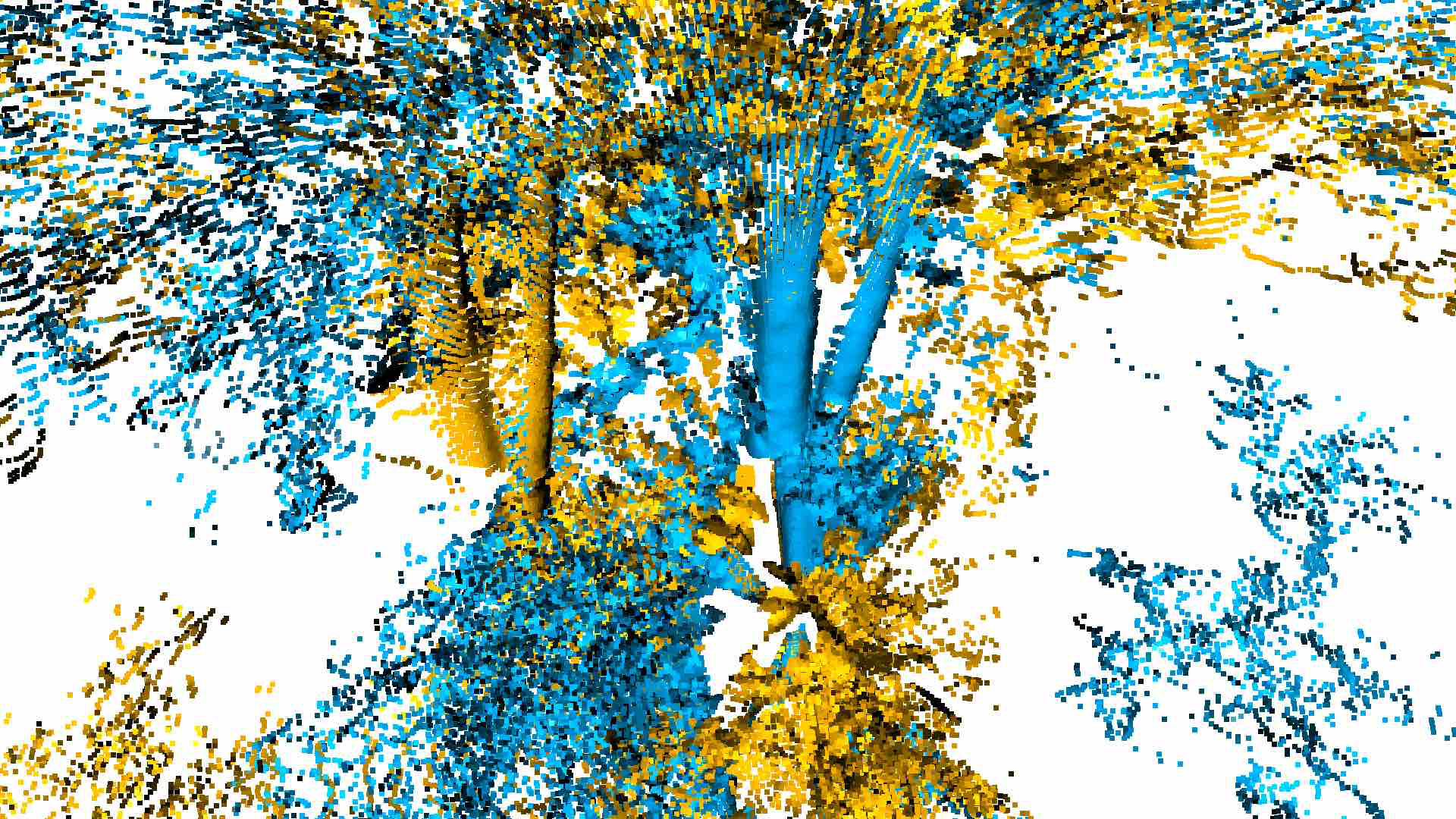}
        \put(25,20){
            \begin{tikzpicture}
            \draw[black, very thick] (0,0) rectangle (0.5,1.1);
            \end{tikzpicture}
        }
        \put(46,20){
            \begin{tikzpicture}
            \draw[black, very thick] (0,0) rectangle (0.5,1.1);
            \end{tikzpicture}
        }
        \put(-3,0){
            \begin{tikzpicture}
            \draw[white, fill=white] (0,0) rectangle (.2,2.4);
            \end{tikzpicture}
        }
        \put(-2,17){\rotatebox{90}{\color{teal}\scriptsize\textbf{correct}}}
        \put(-2,1){\color{black}\scriptsize\textbf{a)}}
        \put(25,59){\color{black}\scriptsize\textbf{before registration}}
    \end{overpic}&
    \begin{overpic}[width=.48\columnwidth]{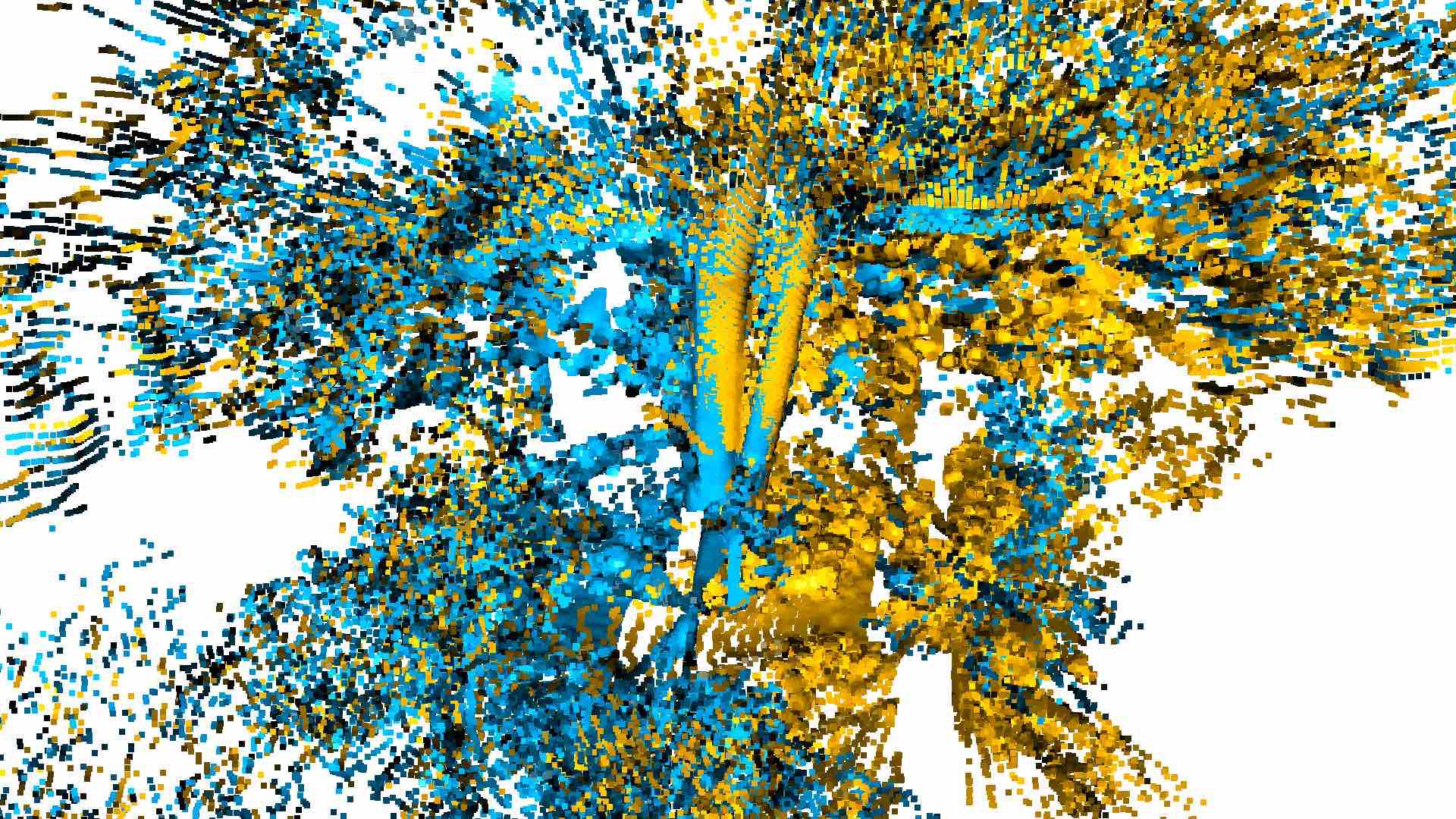}
        \put(42,18){
            \begin{tikzpicture}
            \draw[black, very thick] (0,0) rectangle (0.5,1.2);
            \end{tikzpicture}
        }
        \put(25,59){\color{black}\scriptsize\textbf{after registration}}
    \end{overpic}\\
    \begin{overpic}[width=.48\columnwidth]{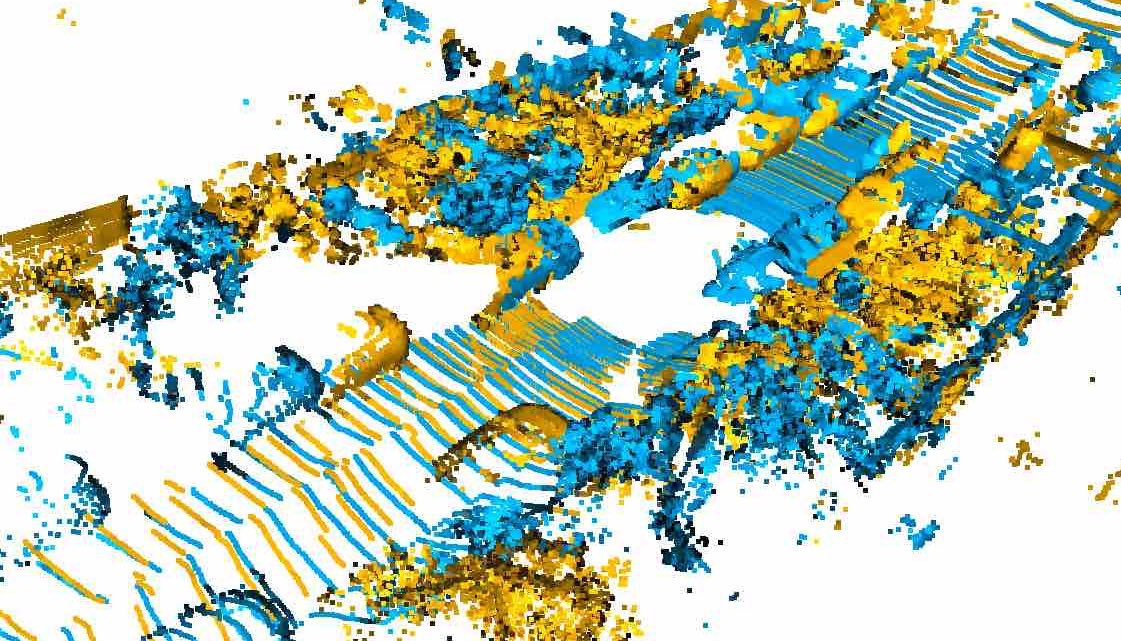}
        \put(15,15){
            \begin{tikzpicture}
            \draw[black, very thick] (0,0) rectangle (0.5,0.5);
            \end{tikzpicture}
        }
        \put(37,26){
            \begin{tikzpicture}
            \draw[black, very thick] (0,0) rectangle (0.5,0.5);
            \end{tikzpicture}
        }
        \put(-3,0){
            \begin{tikzpicture}
            \draw[white, fill=white] (0,0) rectangle (.2,2.4);
            \end{tikzpicture}
        }
        \put(-2,17){\rotatebox{90}{\color{teal}\scriptsize\textbf{correct}}}
        \put(-2,1){\color{black}\scriptsize\textbf{b)}}
    \end{overpic}&
    \begin{overpic}[width=.48\columnwidth]{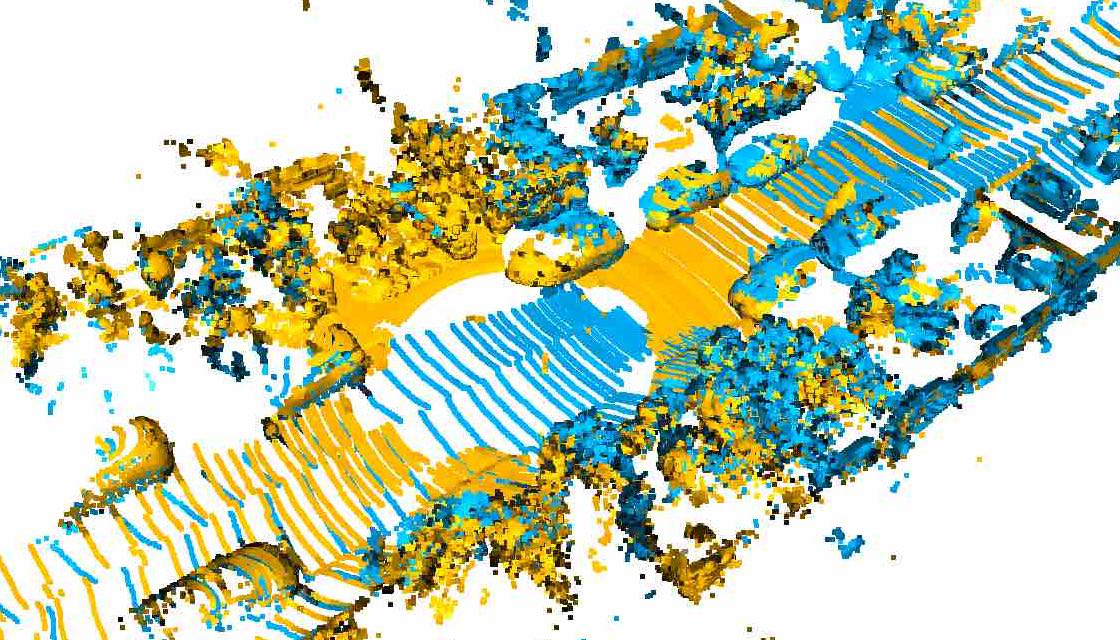}
        \put(20,17){
            \begin{tikzpicture}
            \draw[black, very thick] (0,0) rectangle (0.5,0.5);
            \end{tikzpicture}
        }
    \end{overpic}\\
    \begin{overpic}[width=.48\columnwidth]{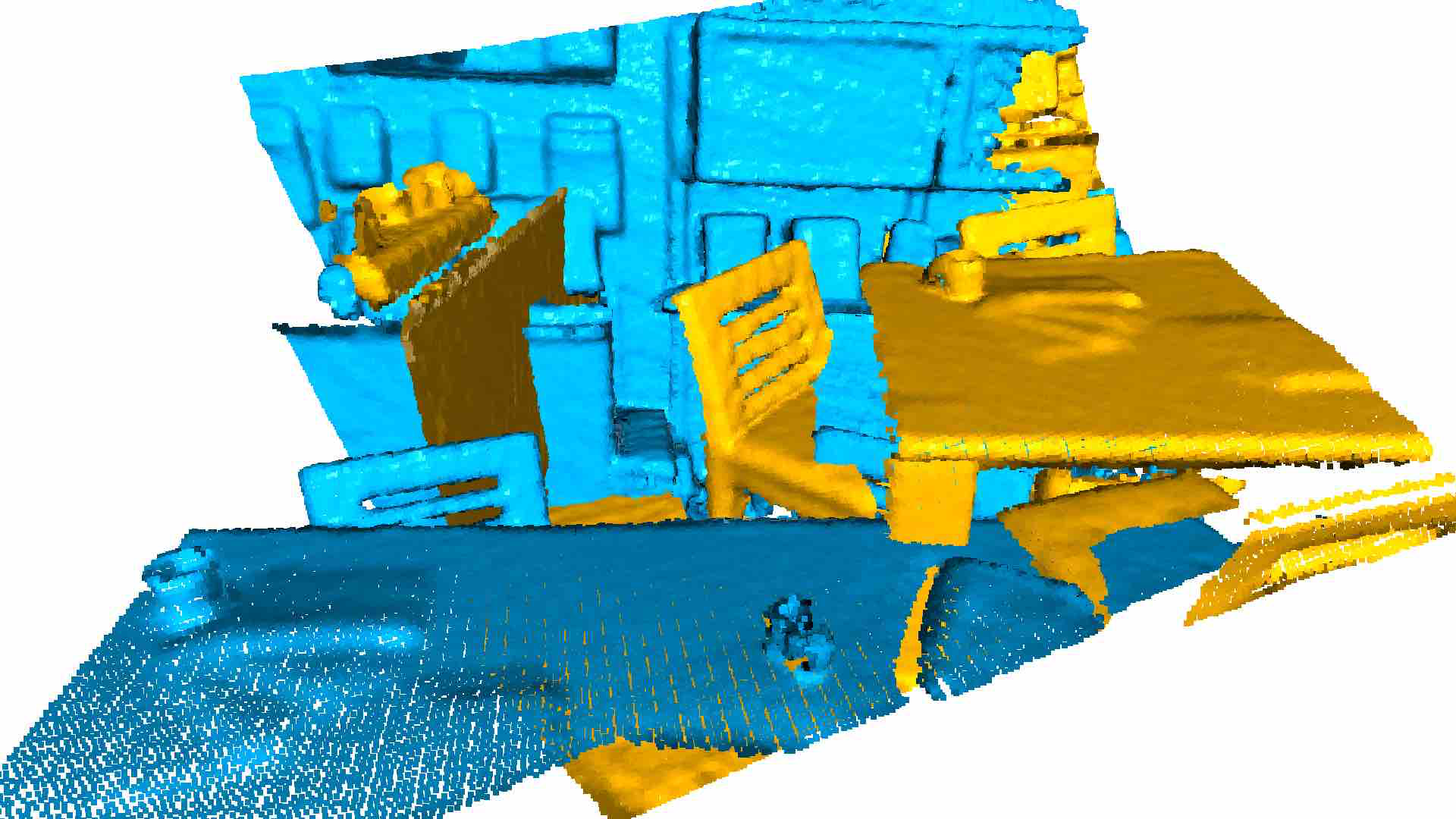}
        \put(5,10){
            \begin{tikzpicture}
            \draw[black, very thick] (0,0) rectangle (0.4,0.4);
            \end{tikzpicture}
        }
        \put(60,33){
            \begin{tikzpicture}
            \draw[black, very thick] (0,0) rectangle (0.3,0.3);
            \end{tikzpicture}
        }
        \put(-3,0){
            \begin{tikzpicture}
            \draw[white, fill=white] (0,0) rectangle (.2,2.4);
            \end{tikzpicture}
        }
        \put(-2,17){\rotatebox{90}{\color{teal}\scriptsize\textbf{correct}}}
        \put(-2,1){\color{black}\scriptsize\textbf{c)}}
    \end{overpic}&
    \begin{overpic}[width=.48\columnwidth]{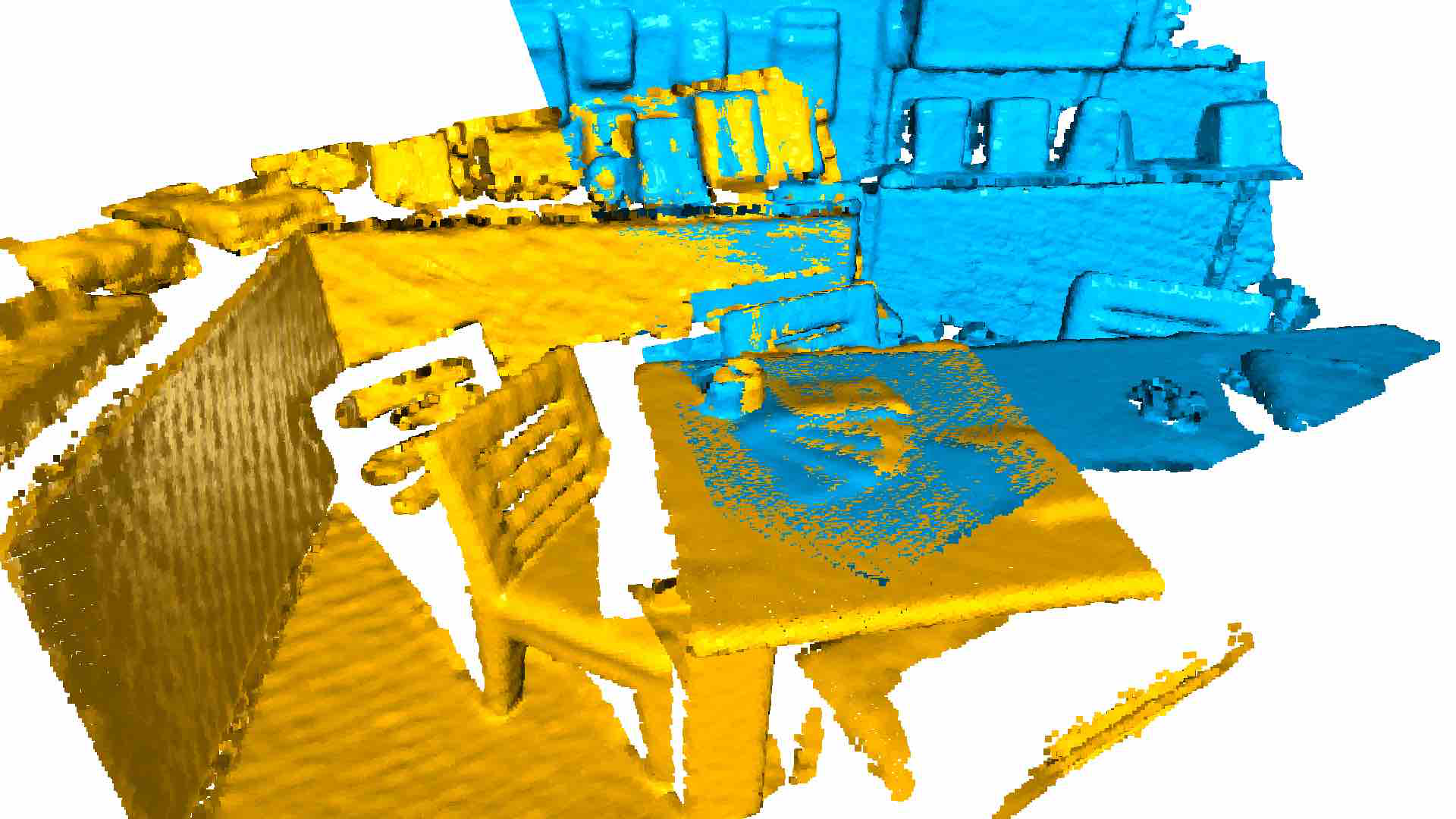}
        \put(43.5,25.25){
            \begin{tikzpicture}
            \draw[black, very thick] (0,0) rectangle (0.3,0.3);
            \end{tikzpicture}
        }
    \end{overpic}\\
    % incorrect
    \begin{overpic}[width=.48\columnwidth]{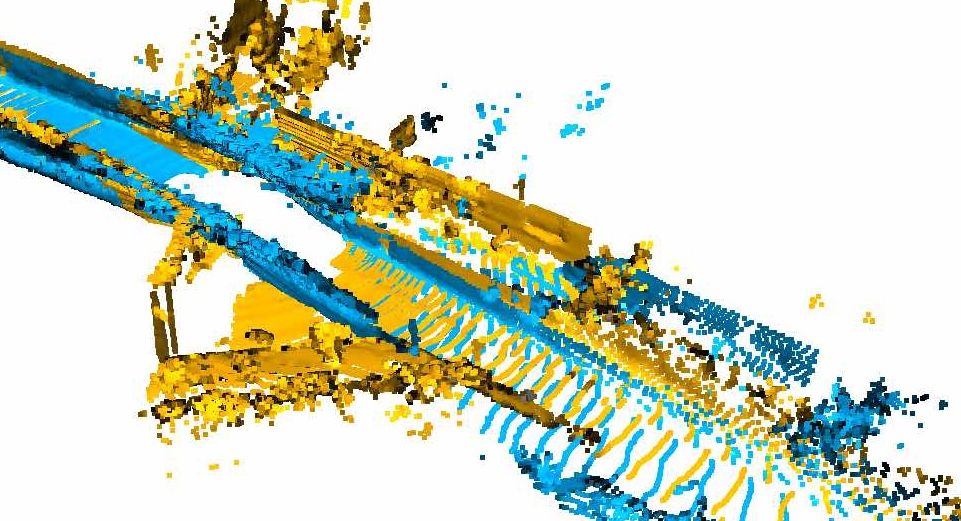}
        \put(54,17){
            \begin{tikzpicture}
            \draw[black, very thick] (0,0) rectangle (0.6,0.6);
            \end{tikzpicture}
        }
        \put(78,3){
            \begin{tikzpicture}
            \draw[black, very thick] (0,0) rectangle (0.6,0.6);
            \end{tikzpicture}
        }
        \put(-3,0){
            \begin{tikzpicture}
            \draw[white, fill=white] (0,0) rectangle (.2,2.4);
            \end{tikzpicture}
        }
        \put(-2,15){\rotatebox{90}{\color{red}\scriptsize\textbf{incorrect}}}
        \put(-2,1){\color{black}\scriptsize\textbf{d)}}
    \end{overpic}&
    \begin{overpic}[width=.48\columnwidth]{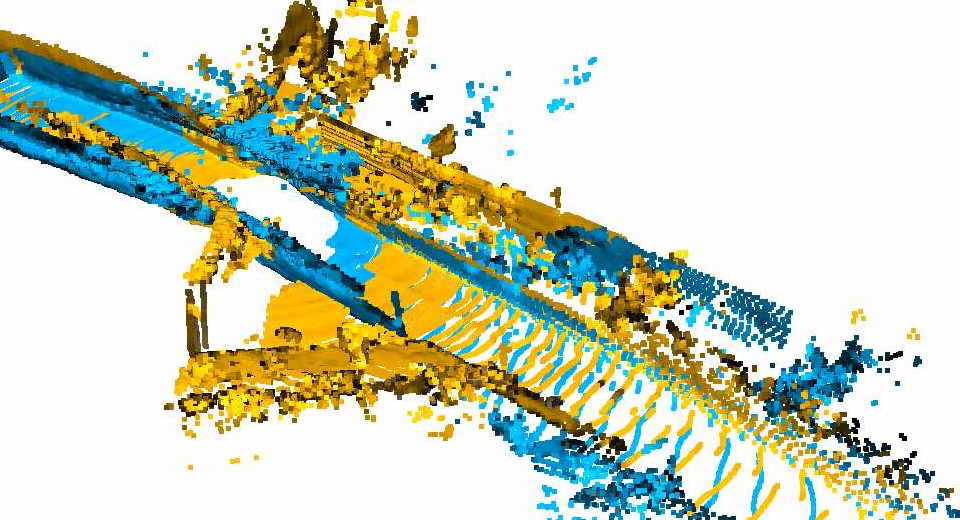}
        \put(57,16){
            \begin{tikzpicture}
            \draw[black, very thick] (0,0) rectangle (0.6,0.6);
            \end{tikzpicture}
        }
        \put(75,7){
            \begin{tikzpicture}
            \draw[black, very thick] (0,0) rectangle (0.6,0.6);
            \end{tikzpicture}
        }
    \end{overpic}\\
    \begin{overpic}[width=.48\columnwidth]{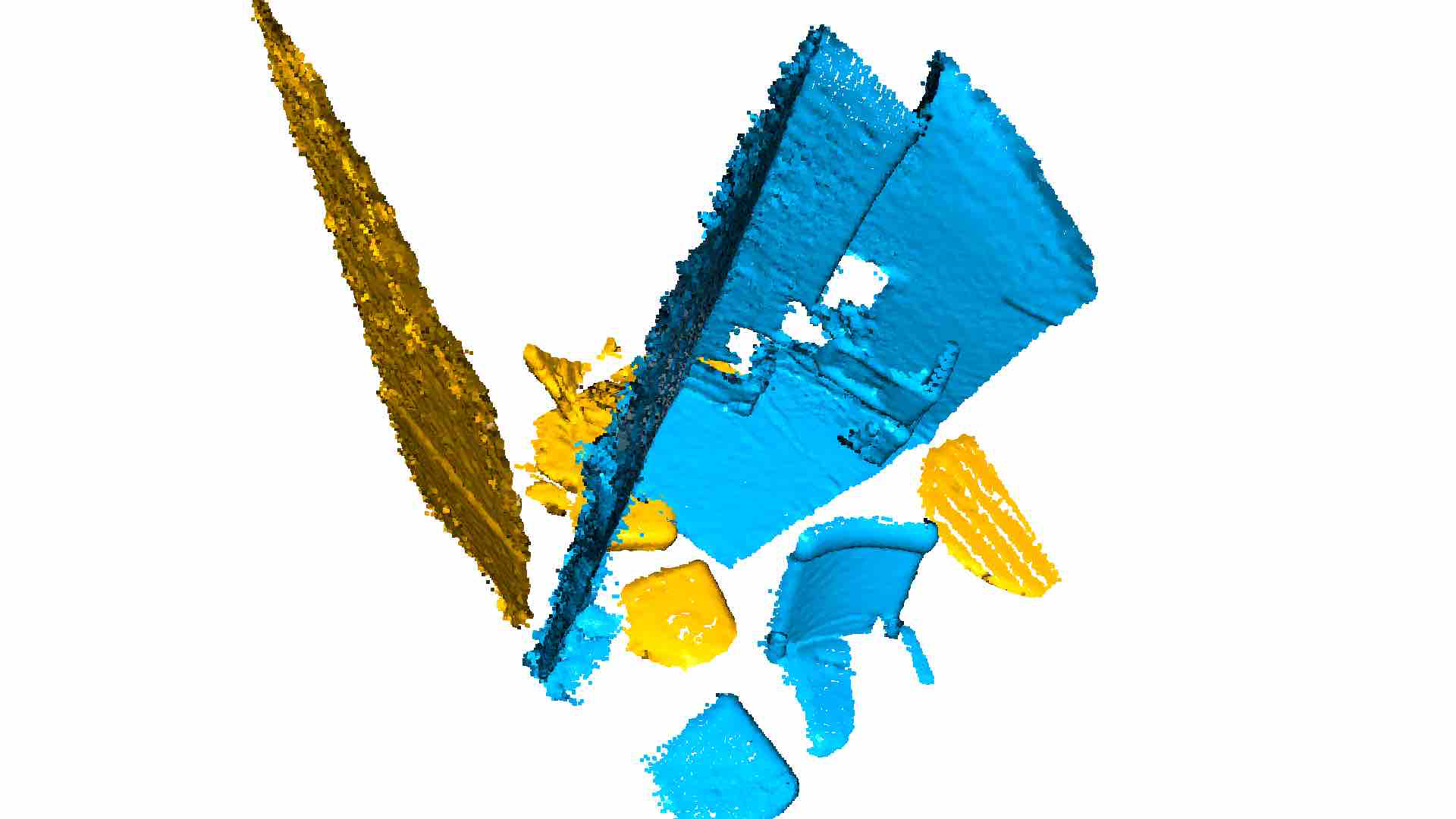}
        \put(49,2){
            \begin{tikzpicture}
            \draw[black, very thick] (0,0) rectangle (0.6,0.8);
            \end{tikzpicture}
        }
        \put(32,20){
            \begin{tikzpicture}
            \draw[black, very thick] (0,0) rectangle (0.4,0.55);
            \end{tikzpicture}
        }
        \put(-2,15){\rotatebox{90}{\color{red}\scriptsize\textbf{incorrect}}}
        \put(-2,1){\color{black}\scriptsize\textbf{e)}}
    \end{overpic}&
    \begin{overpic}[width=.48\columnwidth]{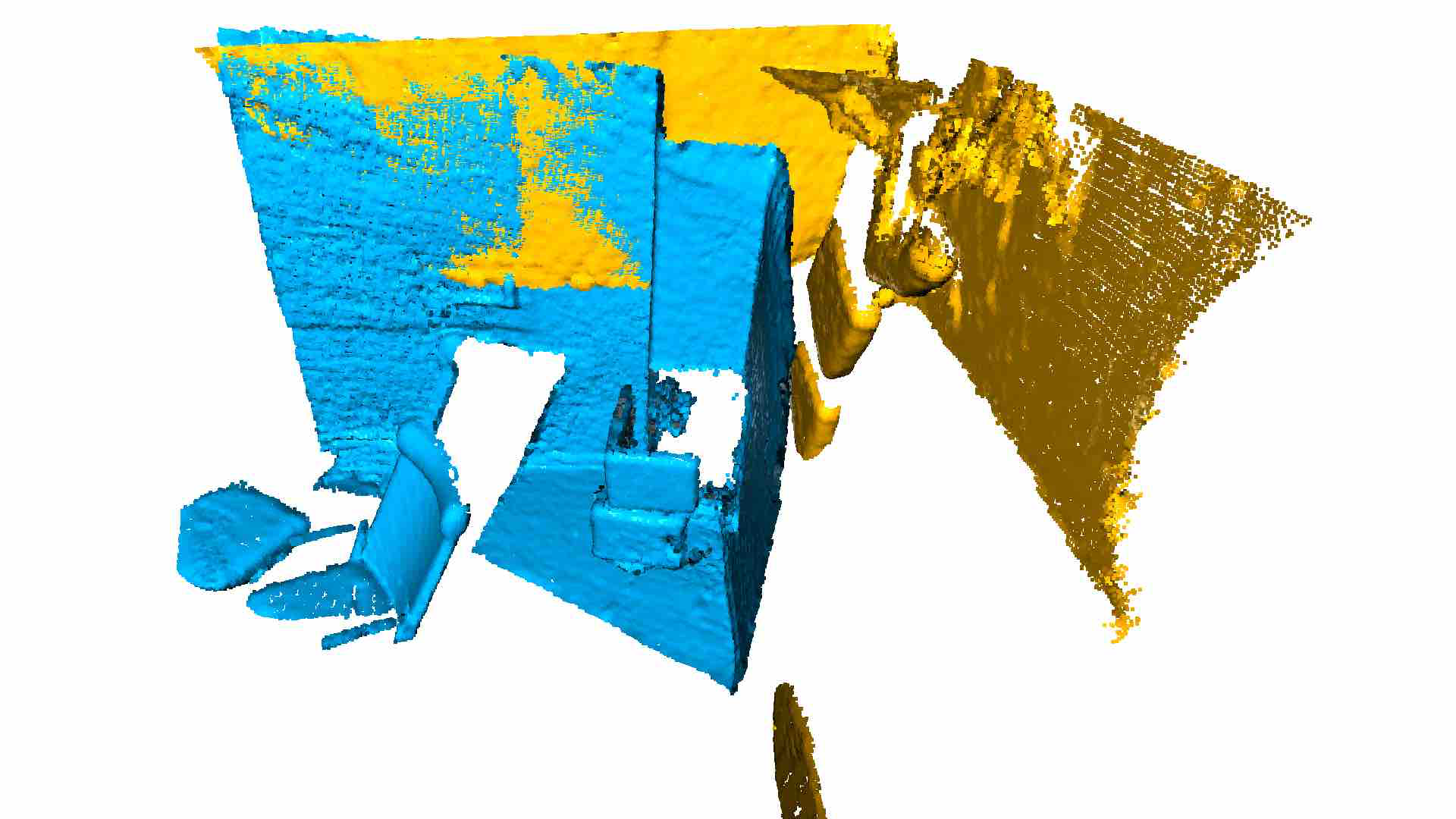}
        \put(49,33.5){
            \begin{tikzpicture}
            \draw[black, very thick] (0,0) rectangle (0.6,0.8);
            \end{tikzpicture}
        }
        \put(20,10){
            \begin{tikzpicture}
            \draw[black, very thick] (0,0) rectangle (0.4,0.7);
            \end{tikzpicture}
        }
    \end{overpic}\\
  \end{tabular}
\end{center}
\vspace{-3mm}
\caption{Qualitative registration results on (a) ETH, (b,d) KITTI and (c,e) 3DMatch datasets using GeDi descriptors trained on 3DMatch. The point clouds are in their original reference frame \emph{before registration} (left-hand side column) and in a common reference frame \emph{after registration} (right-hand side column). We show \emph{correct} and \emph{incorrect} registration results, and highlight regions of interest to facilitate the analysis of the results.}
\label{fig:qualitative_res}
\end{figure}
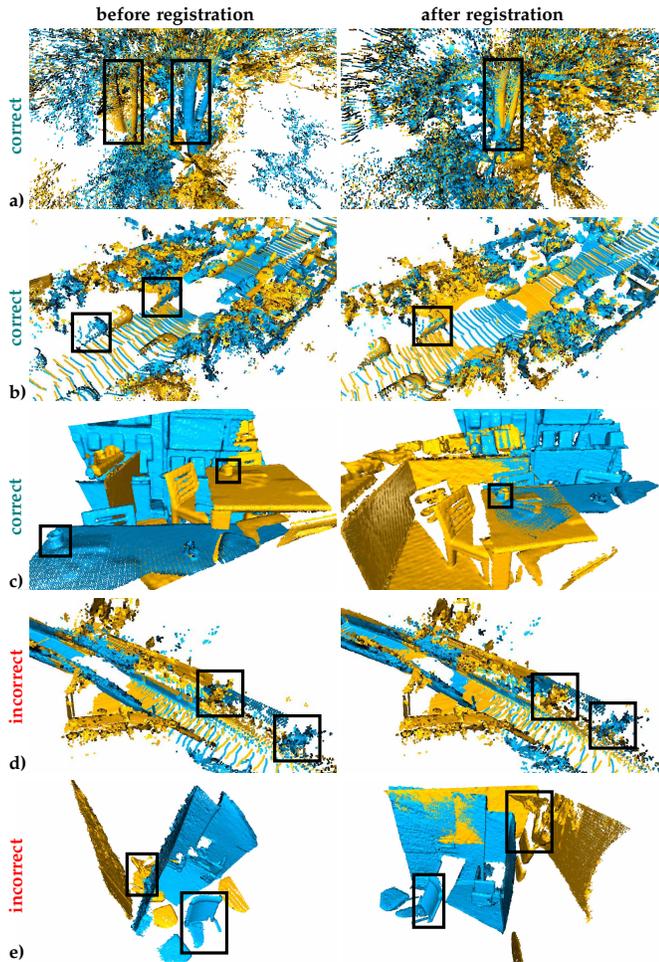

Fig.~\ref{fig:qualitative_res} shows qualitative registration results computed on ETH, KITTI and 3DMatch datasets using GeDi descriptors trained on 3DMatch.
The registration is performed using RANSAC \cite{Zhou2018}.
Figs.~\ref{fig:qualitative_res}a-c are correct registration results.
Although GeDi is trained using data of 3DMatch indoor scenes (c), we can see how GeDi can generalise to the outdoor scenes of ETH (a) and KITTI (b).
The result in (c) is a case with small overlap where we can notice that the registration is correct but not highly accurate.
The point cloud pair in (d) is the incorrect registration result from KITTI discussed in Sec.~\ref{sec:quantitative_analysis}.
This pair is challenging because the overlap region between the two point clouds contains partial structures with little geometric information.
In (e), we can observe that if there are flat surfaces and partially captured objects in the overlap region, the registration may fail.

% ********************************
% ********************************

%%%%%%%%%%%%%%%%%%%%%%%%%%%%%%%%%%%%%%%%%%%%%%%%%%%%%%%%%%%%%%%%%%%%%%
%%%%%%%%%%%%%%%%%%%%%%%%%%%%%%%%%%%%%%%%%%%%%%%%%%%%%%%%%%%%%%%%%%%%%%
%%%%%%%%%%%%%%%%%%%%%%%%%%%%%%%%%%%%%%%%%%%%%%%%%%%%%%%%%%%%%%%%%%%%%%
\section{Conclusions}\label{sec:conclusions}

We presented a novel approach to train general and distinctive 3D local descriptors through a deep network that is invariant to the permutations of the input points and that processes canonicalised patches.
We embedded a quaternion transformation module in our deep network to facilitate patch canonicalisation, which is trained concurrently with the encoding module.
Distinctiveness is achieved by training our deep network with a Siamese approach by using a hardest-contrastive loss.
Compared to existing deep learning-based 3D descriptors, GeDi can better generalise across domains, while also being as effective as the other descriptors when it is trained and tested in the same domain.
Future research directions include building end-to-end trainable LRF-based descriptors, designing a strategy to learn patch radii and improving computational efficiency.

\bibliographystyle{IEEEtran}
\bibliography{refs}

\begin{thebibliography}{10}
\providecommand{\url}[1]{#1}
\csname url@samestyle\endcsname
\providecommand{\newblock}{\relax}
\providecommand{\bibinfo}[2]{#2}
\providecommand{\BIBentrySTDinterwordspacing}{\spaceskip=0pt\relax}
\providecommand{\BIBentryALTinterwordstretchfactor}{4}
\providecommand{\BIBentryALTinterwordspacing}{\spaceskip=\fontdimen2\font plus
\BIBentryALTinterwordstretchfactor\fontdimen3\font minus
  \fontdimen4\font\relax}
\providecommand{\BIBforeignlanguage}[2]{{%
\expandafter\ifx\csname l@#1\endcsname\relax
\typeout{** WARNING: IEEEtran.bst: No hyphenation pattern has been}%
\typeout{** loaded for the language `#1'. Using the pattern for}%
\typeout{** the default language instead.}%
\else
\language=\csname l@#1\endcsname
\fi
#2}}
\providecommand{\BIBdecl}{\relax}
\BIBdecl

\bibitem{Zhou2022}
Y.~Zhou, Y.~Wang, F.~Poiesi, Q.~Qin, and Y.~Wan, ``{Loop closure detection
  using local 3D deep descriptors},'' \emph{RAL}, vol.~7, no.~3, pp.
  6335--6342, Jul 2022.

\bibitem{Drost2010}
B.~Drost, M.~Ulrich, N.~Navab, and S.~Ilic, ``{Model Globally, Match Locally:
  Efficient and Robust 3D Object Recognition},'' in \emph{CVPR}, 2010.

\bibitem{Rusu2009}
R.~Rusu, N.~Blodow, and M.~Beetz, ``{Fast point feature histograms for 3D
  registration},'' in \emph{ICRA}, 2009.

\bibitem{Deng2018cvpr}
H.~Deng, T.~Birdal, and S.~Ilic, ``{PPFNet: Global context aware local features
  for robust 3D point matching},'' in \emph{CVPR}, 2018.

\bibitem{Poiesi2021}
F.~Poiesi and D.~Boscaini, ``{Distinctive 3D local deep descriptors},'' in
  \emph{ICPR}, 2021.

\bibitem{Ao2021}
S.~Ao, Q.~Hu, B.~Yang, A.~Markham, and Y.~Guo, ``{SpinNet: Learning a General
  Surface Descriptor for 3D Point Cloud Registration},'' in \emph{CVPR}, 2021.

\bibitem{Besl1992}
P.~Besl and N.~McKay, ``{A method for registration of 3-D shapes},''
  \emph{PAMI}, vol.~14, no.~2, pp. 239--256, Feb 1992.

\bibitem{Fischler1981}
M.~Fischler and R.~Bolles, ``Random sample consensus: A paradigm for model
  fitting with applications to image analysis and automated cartography,''
  \emph{Commun. ACM}, vol.~24, no.~6, pp. 341--406, Jun 1981.

\bibitem{Zeng2017}
A.~Zeng, S.~Song, M.~Niessner, M.~Fisher, J.~Xiao, and T.~Funkhouser,
  ``{3DMatch: Learning the matching of local 3D geometry in range scans},'' in
  \emph{CVPR}, 2017.

\bibitem{Choy2019}
C.~Choy, J.~Park, and V.~Koltun, ``{Fully Convolutional Geometric Features},''
  in \emph{ICCV}, 2019.

\bibitem{Bai2020}
X.~Bai, Z.~Luo, L.~Zhou, H.~Fu, L.~Quan, and C.-L. Tai, ``{D3Feat: Joint
  Learning of Dense Detection and Description of 3D Local Features},'' in
  \emph{CVPR}, Jun 2020.

\bibitem{Geiger2012}
A.~Geiger, P.~Lenz, and R.~Urtasun, ``{Are we ready for autonomous driving? The
  KITTI vision benchmark suite},'' in \emph{CVPR}, Jun 2012.

\bibitem{Salti2014}
S.~Salti, F.~Tombari, and L.~Di~Stefano, ``{SHOT: Unique signatures of
  histograms for surface and texture description},'' \emph{CVIU}, vol. 125, pp.
  251--264, Aug 2014.

\bibitem{Yang2016}
J.~Yang, Q.~Zhang, Y.~Xiao, and Z.-G. Cao, ``{TOLDI: An effective and robust
  approach for 3D local shape description},'' \emph{Patt. Recogn.}, vol.~65,
  pp. 175--187, May 2017.

\bibitem{Gojcic2019}
Z.~Gojcic, C.~Zhou, J.~Wegner, and W.~Andreas, ``{The perfect match: 3D point
  cloud matching with smoothed densities},'' in \emph{CVPR}, 2019.

\bibitem{Deng2018eccv}
H.~Deng, T.~Birdal, and S.~Ilic, ``{PPF-FoldNet: Unsupervised learning of
  rotation invariant 3D local descriptors},'' in \emph{ECCV}, 2018.

\bibitem{Deng2019}
------, ``{3D local features for direct pairwise registration},'' in
  \emph{CVPR}, 2019.

\bibitem{Zhao2019}
Y.~Zhao, T.~Birdal, H.~Deng, and F.~Tombari, ``{3D point capsule networks},''
  in \emph{CVPR}, 2019.

\bibitem{Johnson1999}
A.~Johnson and M.~Hebert, ``{Using spin images for efficient object recognition
  in cluttered 3D scenes},'' \emph{PAMI}, vol.~21, no.~5, pp. 433--449, May
  1999.

\bibitem{Zhong2009}
Y.~Zhong, ``{Intrinsic shape signatures: A shape descriptor for 3D object
  recognition},'' in \emph{ICCV}, 2009.

\bibitem{Qi2017b}
C.~Qi, L.~Yi, H.~Su, and L.~Guibas, ``{PointNet++: Deep Hierarchical Feature
  Learning on Point Sets in a Metric Space},'' in \emph{NIPS}, 2017.

\bibitem{Pomerleau2012}
F.~Pomerleau, M.~Liu, F.~Colas, and R.~Siegwart, ``{Challenging data sets for
  point cloud registration algorithms},'' \emph{IJRR}, vol.~31, no.~14, pp.
  1705--1711, Dec 2012.

\bibitem{Cirujeda2014}
P.~Cirujeda, X.~Mateo, Y.~Cid, and X.~Binefa, ``{MCOV: a covariance descriptor
  for fusion of texture and shape features in 3D point clouds},'' in
  \emph{3DV}, 2014.

\bibitem{Guo2013}
Y.~Guo, F.~Sohel, M.~Bennamoun, M.~Lu, and J.~Wan, ``{Rotational projection
  statistics for 3D local surface description and object recognition},''
  \emph{IJCV}, vol. 105, pp. 63--86, Apr 2013.

\bibitem{Qi2017a}
C.~Qi, H.~Su, K.~Mo, and L.~Guibas, ``{PointNet: Deep learning on point sets
  for 3D classification and segmentation},'' in \emph{CVPR}, 2017.

\bibitem{Yang2018}
Y.~Yang, C.~F.~Y. Shen, and D.~Tian, ``{FoldingNet: Point cloud auto-encoder
  via deep grid deformation},'' in \emph{CVPR}, 2018.

\bibitem{Thomas2017}
H.~Thomas, C.~Qi, J.-E. Deschaud, B.~Marcotegui, F.~Goulette, and L.~Guibas,
  ``{KPConv: Flexible and deformable convolution for point clouds},'' in
  \emph{ICCV}, 2017.

\bibitem{Li2020}
L.~Li, S.~Zhu, H.~Fu, P.~Tan, and C.-L. Tai, ``{End-to-End Learning Local
  Multi-view Descriptors for 3D Point Clouds},'' in \emph{CVPR}, 2020.

\bibitem{Spezialetti2019}
R.~Spezialetti, S.~Salti, and L.~D. Stefano, ``{Learning an Effective
  Equivariant 3D Descriptor Without Supervision },'' in \emph{ICCV}, 2019.

\bibitem{Yuan2019}
W.~Yuan, D.~Held, C.~Mertz, and M.~Hebert, ``{Iterative Transformer Network for
  3D Point Cloud},'' in \emph{arXiv:1811.11209}, 2018.

\bibitem{Tombari2010}
F.~Tombari, S.~Salti, and L.~D. Stefano, ``{Unique shape context for 3D data
  description},'' in \emph{ACM 3D Object Retrieval}, 2010.

\bibitem{Khoury2017}
M.~Khoury, Q.-Y. Zhou, and V.~Koltun, ``{Learning compact geometric
  features},'' in \emph{ICCV}, 2017.

\bibitem{Yew2018}
Z.~Yew and G.~Lee, ``{3DFeat-Net: Weakly supervised local 3D features for point
  cloud registration},'' in \emph{ECCV}, 2018.

\bibitem{Ma2016}
Y.~Ma, Y.~Guo, J.~Zhao, M.~Lu, J.~Zhang, and J.~Wan, ``{Fast and accurate
  registration of structured point clouds with small overlaps},'' in \emph{CVPR
  Workshops}, 2016.

\bibitem{Zhou2018}
Q.-Y. Zhou, J.~Park, and V.~Koltun, ``{Open3D}: {A} modern library for {3D}
  data processing,'' \emph{arXiv:1801.09847}, 2018.

\end{thebibliography}

\vfill

\end{document}